\documentclass[lettersize,journal]{IEEEtran}
\usepackage{amsmath,amsfonts}
\usepackage{algorithmic}
\usepackage{algorithm}

\usepackage{array}
\usepackage[caption=false,font=normalsize,labelfont=sf,textfont=sf]{subfig}
\usepackage{textcomp}
\usepackage{stfloats}
\usepackage{url}
\usepackage{verbatim}
\usepackage{graphicx}
\usepackage{cite}
\usepackage{hyperref}
\usepackage[font=scriptsize,labelfont=scriptsize]{caption}
\hypersetup{
    colorlinks=true,
    linkcolor=blue, 
    citecolor=blue,
    urlcolor=blue
}

\newcommand{\figref}[1]{%
    \hyperref[#1]{Fig.~\ref*{#1}}%
}
\newcommand{\algoref}[2][]{%
    \hyperref[#2]{#1}%
}
\newcommand{\secref}[1]{%
    \hyperref[#1]{Sec.~\ref*{#1}}%
}

\newcommand{\tableref}[1]{%
    \hyperref[#1]{Tab.~\ref*{#1}}%
}
\newcommand{\eqrefp}[1]{%
    \hyperref[#1]{Eq.~\ref*{#1}}%
}
\usepackage{float}

\usepackage{balance}
\begin{document}

\title{Towards Robust and Generalizable Gerchberg Saxton based Physics Inspired Neural Networks for Computer Generated Holography: A Sensitivity Analysis Framework}
\author{Ankit Amrutkar$^{*1,2}$, Björn Kampa$^{2,3}$, Volkmar Schulz$^{1}$, Johannes Stegmaier$^{\ddag 1}$, Markus Rothermel$^{\ddag *4}$, Dorit Merhof$^{\ddag *2,5}$
\thanks{\normalfont{[1]} Institute of Imaging and Computer Vision, RWTH Aachen University, 52062, Aachen, Germany.
\normalfont{[2]} Research Training Group 2416 MultiSenses MultiScales, RWTH Aachen University, 52074 Aachen, Germany.
\normalfont{[3]} Department of Molecular and Systemic Neurophysiology, Institute of Biology II, RWTH Aachen University, 52074  Aachen
Germany.
\normalfont{[4]} Institute of Physiology, RG Neurophysiology and Optogenetics, Medical Faculty, Otto-von-Guericke-University, Magdeburg, Germany.
\normalfont{[5]} Image Analysis and Computer Vision, Regensburg University, 93040 Regensburg, Germany. 
This work was funded by the Deutsche Forschungsgemeinschaft (DFG, German Research Foundation) - 368482240/GRK2416. $^{\ddag}$ Equal contributions. $\textit{* Corresponding authors }$(ankit.amrutkar@lfb.rwth-aachen.de, markus.rothermel@med.ovgu.de, dorit.merhof@ur.de)}

}



\maketitle

\begin{abstract}
Computer-generated holography (CGH) enables applications in holographic augmented reality (AR), 3D displays, systems neuroscience, and optical trapping. The fundamental challenge in CGH is solving the inverse problem of phase retrieval from intensity measurements. Physics-inspired neural networks (PINNs), especially Gerchberg-Saxton-based PINNs (GS-PINNs), have advanced phase retrieval capabilities. However, their performance strongly depends on forward models (FMs) and their hyperparameters (FMHs), limiting generalization, complicating benchmarking, and hindering hardware optimization. 
We present a systematic sensitivity analysis framework based on Saltelli's extension of Sobol's method to quantify FMH impacts on GS-PINN performance. Our analysis demonstrates that SLM pixel-resolution is the primary factor affecting neural network sensitivity, followed by pixel-pitch, propagation distance, and wavelength. Free space propagation forward models demonstrate superior neural network performance compared to Fourier holography, providing enhanced parameterization and generalization. We introduce a composite evaluation metric combining performance consistency, generalization capability, and hyperparameter perturbation resilience, establishing a unified benchmarking standard across CGH configurations.
Our research connects physics-inspired deep learning theory with practical CGH implementations through concrete guidelines for forward model selection, neural network architecture, and performance evaluation. Our contributions advance the development of robust, interpretable, and generalizable neural networks for diverse holographic applications, supporting evidence-based decisions in CGH research and implementation.
\end{abstract}


\section{Introduction}
\begin{figure}[H]
	\centering
	\includegraphics[width=3.5in]{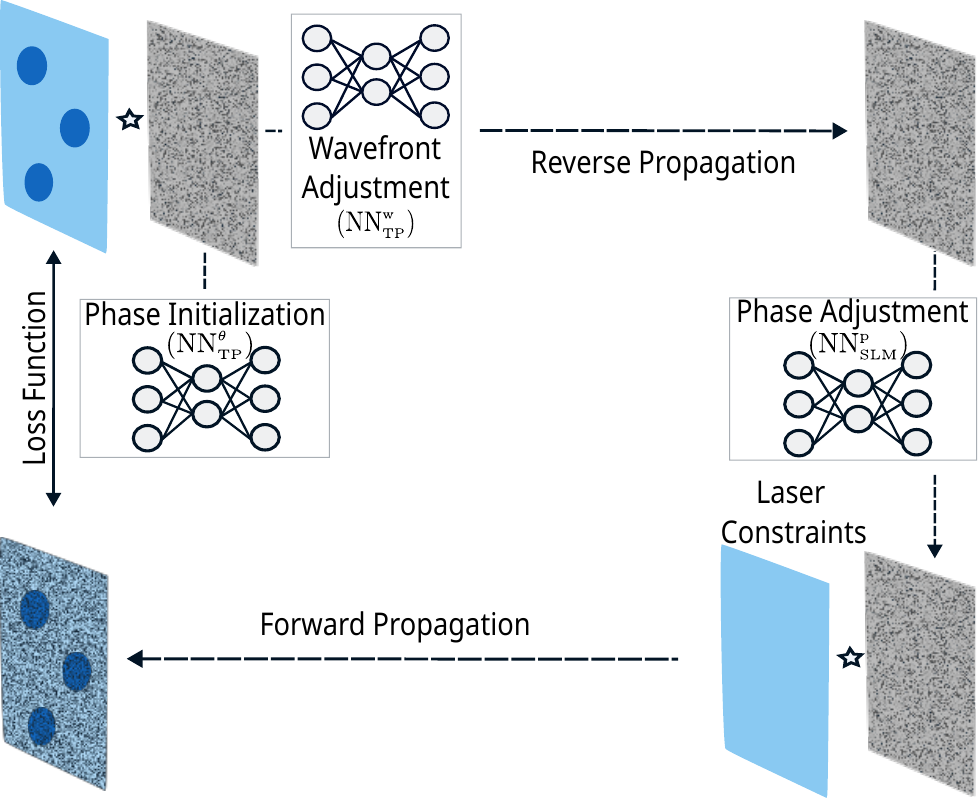}
	\caption{GS-Physics Inspired Neural Network (GS-PINN) with Phase Initialization, Wavefront and Phase Adjustment Neural networks (\algoref[Algorithm 2]{algorithm:unrolled_GS_PINN}). The laser constraints consist of a linearly polarized beam with uniform amplitude, and the 'star' symbol represents the formation of a complex wavefront.}
	\label{fig:GS_PINN}
\end{figure}

\IEEEPARstart{C}{omputer}-generated holography (CGH), is a technique used to create specific light intensity patterns by controlling a coherent light wave. This is usually achieved by digitally adjusting the wave's phase using a spatial light modulator (SLM). CGH algorithms determine the optimal way to modulate a wave by solving a complex inverse problem that is ill-posed, nonlinear, and non-convex. CGH, a key area in computational imaging has a range of applications in holographic augmented reality, 3D displays \cite{He:19,Blinder2022,pi2022review}, systems neuroscience \cite{marshel2019cortical,emiliani2015all,pegard2017three,Eybposh:2022}, optical trapping \cite{keir2004optical,curtis2002dynamic,leach20043d,favre2022optical}, and more. Various deep learning-based methods \cite{Barbastathis:19,Zhangyixin2022,Situ2022,ciarlo2024deep,ren2024creation} exist to solve the problem faster than traditional iterative methods\cite{MADSEN2023129458}, such as the Gerchberg-Saxton (GS) algorithm \cite{1}. One potential approach is to modify iterative methods like the GS algorithm using model-based deep learning techniques\cite{2}, which combine the strong performance of iterative methods with the faster inference times of neural networks. A natural extension of the GS algorithm is its unrolling \cite{9363511}, where neural networks replace the initial conditions of the GS algorithm, the iterative process is guided by a loss function, and additional adjustment neural networks are included at either the image plane or/and at the SLM plane (\algoref[Algorithm 2]{algorithm:unrolled_GS_PINN}). This approach also allows for unsupervised training. Such an unrolling of the GS algorithm can be referred as GS model-based Physics-Inspired Neural Networks (GS-PINN)\cite{schlieder2020learned,3} (\figref{fig:GS_PINN}).

In these approaches, the physics of the forward model is dictated by the hardware configuration. In some holographic Augmented Reality (AR) systems, free space propagation is commonly employed to simulate how light moves from the SLM to the viewer’s eye. One widely used computational technique for this is the Angular Spectrum Method (ASM)\cite{kim2011digital,Shi2021,Zhong2024}. Key hyperparameters for ASM include the wavelength of light, the distance between the SLM and the image plane, the SLM pixel-pitch, and its size (\figref{fig:fm_fmh}). In systems neuroscience, holographic methods are often applied in optogenetics and brain stimulation\cite{Russell2022}, where precise light field control is essential for accurately targeting neurons\cite{papagiakoumou2018two,HosseinEybposh:20}. Across various scientific fields like chemistry \cite{heller2014optical,
kulin2003optical}, material sciences \cite{minowa2022optical}, biophysics \cite{diekmann2016nanoscopy,nussenzveig2018cell} and quantum science \cite{kaufman2021quantum}, holographic optical tweezers (HOTs) \cite{hayasaki1999nonmechanical} utilize CGH \cite{wang2006dynamic} to create arbitrary tweezer geometries, enabling the simultaneous manipulation of multiple particles with enhanced flexibility and control. Fourier holography\cite{SPALDING2008139,Eybposh:2022,Ersaro:23,Madsen2022} is typically utilized in these contexts, with important parameters being the SLM pixel-resolution/size. 

Given the widespread application of CGH across fields ranging from augmented reality and neuroscience to material and quantum sciences/technologies, developing robust, interpretable, and generalizable neural network models for phase retrieval is essential. Interpreting unsupervised CGH networks as unrolled GS algorithm enhances explainability of the models. However, the performance of neural networks in phase retrieval \cite{dong2023phase} is highly sensitive to the choice of forward models (FMs) and their associated hyperparameters (FMHs), requiring different networks to be designed and trained for each new FM-FMH configuration. This sensitivity presents three key challenges. First, network performance varies significantly across FMs and FMHs, making it difficult to develop models that are robust across configurations. Second, models trained on specific FMs and FMHs often fail to generalize, limiting their adaptability for experimentalists and theorists who require versatile and transferable solutions. Third, FMH sensitivity complicates benchmarking, making reliable comparisons between models trained on different configurations challenging. Addressing these issues requires a systematic analysis of how perturbations in FM-FMH configurations influence GS-PINN performance, impact interpretability and generalization, and how benchmarking methodologies can be refined to account for these sensitivities.


To address these challenges, we develop a structured framework that includes sensitivity analysis, forward model evaluation, and benchmarking to improve the interpretability, robustness, and evaluation of neural networks in phase retrieval tasks. Specifically, we introduce a variance-based quasi Monte Carlo approach using Saltelli’s extension of Sobol’s method \cite{saltelli2008global, SALTELLI2010259} to quantify the sensitivity of FMHs on neural network performance. Applied to the GS algorithm and GS-PINN, this analysis identifies the key hyperparameters that influence performance, offering critical insights for network design and optimization.

We also evaluated the effects of different FMs using a general Monte Carlo approach, comparing Fourier holography and free space propagation. For GS-PINN, our analysis demonstrates that free space propagation consistently outperforms Fourier holography, providing experimentalists with a basis for selecting optimal FMs in hardware setups. To further address the challenges of benchmarking, we developed a composite metric to enable standardized evaluation across configurations. While this metric balances the need for consistent comparison with the risk of speculative conclusions, it highlights the limitations inherent in benchmarking networks with differing FM dependencies. Finally, this work enhances the interpretability and generalization of neural networks in phase retrieval. By identifying the FMHs most affecting performance, our approach provides a pathway to designing more explainable AI models. Collectively, our contributions bridge the gap between theoretical models and practical applications, offering tools for designing, evaluating, and understanding neural networks in diverse CGH tasks.

We summarize the contributions of the work:
\begin{enumerate}
	\item Sensitivity analysis of FMH: We introduced a variance-based quasi-Monte Carlo approach to quantify the impact of FMHs (ASM) on neural network performance. This aids in understanding the relationship between FMHs and network performance. We quantified the sensitivity of FMHs for the GS algorithm and GS-PINN.
	\item Forward model sensitivity: We perform a general Monte-Carlo approach to compare the eﬀects of diﬀerent forward models. This aids the experimentalists using neural networks to choose the FM's accordingly.
	\item Benchmarking and evaluation: We addressed the complexity of evaluating diﬀerent networks and developed a composite metric, highlighting its limitations. This metric represents a compromise, eﬀectively balancing the risk of unreliable evaluations with the prevention of speculative interpretations or claims.
	\item Model interpretability and generalization: Our approach identifies key FMHs influencing network performance, improving model interpretability and paves the way for the development of generalized, explainable AI with careful output interpretation.
\end{enumerate}
The paper is organized as follows: \secref{sec:method:preliminaries} covers preliminaries, including the GS algorithm, its unrolling, and forward models. \secref{sec:method:sa_FMH} analyzes FMH sensitivity using Sobol's method, while \secref{sec:method:sa_FM} extends this to forward models. \secref{sec:method:metric} introduces evaluation metrics and their limitations. Results are presented in \secref{sec:results}, followed by discussion and conclusions in \secref{sec:discussion} and \secref{sec:conclusion}.

\section{Methods}
\label{sec:Method}
\subsection{Preliminaries}
\label{sec:method:preliminaries}
\subsubsection{Gerchberg-Saxton (GS) Algorithm}
The GS algorithm is an iterative phase retrieval algorithm. It approximates the phase at a given plane from its intensity measurement. This is achieved by propagating the wavefront of light between two different planes\cite{1}. The hardware setup between the two planes determines which forward models are used to evaluate the propagation of light. In the context of computer generated holography, one plane is the hologram plane and the other is the SLM plane. Here the phase at the SLM plane is approximated to generate a hologram of a known intensity pattern. The pseudocode is shown in \algoref[Algorithm 1]{algorithm:GS_Algorithm}.

\begin{figure}
	\centering
        \includegraphics[height=5.7in]{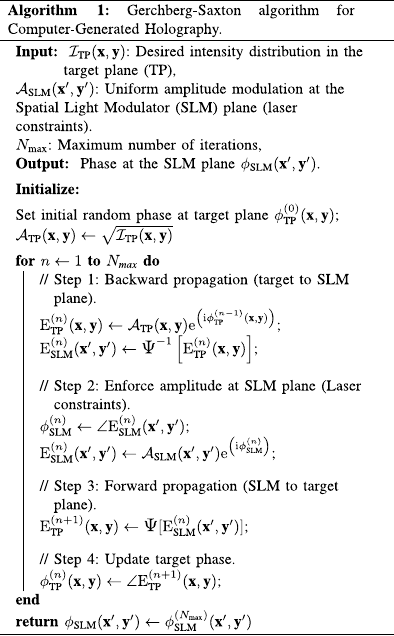}
	\caption*{}
	\label{algorithm:GS_Algorithm}
\end{figure}
\begin{figure} 
	\centering
	\includegraphics[width=3.0in]{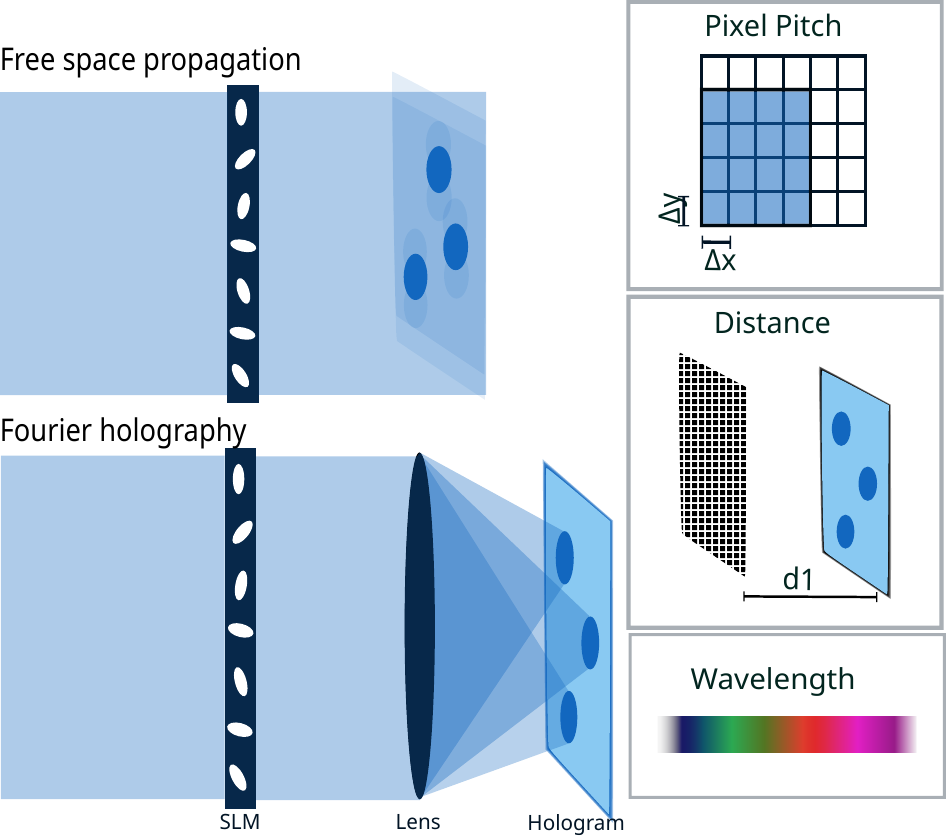}
	\caption{Forward Models (FM) and Forward Model Hyperparameters (FMH). For Fourier holography SLM pixel-resolution is the FMH (\eqrefp{eq:Fourier_forward_model}). For free space propagation wavelength of light, propagation distance, SLM pixel-resolution and pixel-pitch are the FMH (\eqrefp{eq:ASM_forward_model}).  }
	\label{fig:fm_fmh}
\end{figure}

\subsubsection{Forward models}
Here we discuss two forward models \cite{kim2011digital,goodman2005introduction,Matsushima2020} (\figref{fig:fm_fmh}). 
\paragraph{Fourier Holography}
In Fourier holography \cite{goodman2005introduction}, \figref{fig:fm_fmh}, \eqrefp{eq:Fourier_forward_model} there is a lens in between the SLM and hologram plane located at the front and back focal plane of the lens. $\Gamma(\textbf{x},\textbf{y}), \Gamma(\textbf{u},\textbf{v})$ is the wavefield at the front and back focal plane of the lens. $\gamma$ is a phase factor $\exp(2\mathrm{i}kf)/\mathrm{i}\lambda f$, where $k,\lambda, f$ are wave number, wavelength, and focal length of the lens. $\mathcal{F}$ is the Fourier transform and $\widetilde{\mathcal{F}}$ is the inverse Fourier transform.

\begin{equation}
	\label{eq:Fourier_forward_model}
	\begin{aligned}
		\scalebox{1.2}{$\Psi$}_{\text{Fourier}}\left(\Gamma(\textbf{x},\textbf{y})\right)      & =\gamma\mathcal{F}\left(\Gamma(\textbf{x},\textbf{y})\right) & =\Gamma(\textbf{u},\textbf{v}) \\
		\scalebox{1.2}{$\Psi$}_{\text{Fourier}}^{-1}\left(\Gamma(\textbf{u},\textbf{v})\right) & =
		\gamma \widetilde{\mathcal{F}}\left(\Gamma(\textbf{u},\textbf{v})\right)                            & =\Gamma(\textbf{x},\textbf{y})
	\end{aligned}
\end{equation}
\paragraph{Free space propagation}
Here there is no diffractive element in between the SLM and the hologram plane. We use the band limited angular spectrum method \cite{Matsushima:09,Matsushima2020} to simulate the forward and reverse propagation of the light. Here the FMHs are wavelength of light ($\lambda$), propagation distance ($d$), SLM pixel-resolution ($M$) and pixel-pitch ($\Delta x$) (\eqrefp{eq:ASM_forward_model}). 
\begin{equation}
	\label{eq:ASM_forward_model}
	\begin{aligned}
		\text{FMH}                                                       & \equiv \left(\lambda,\Delta x,M,d\right)                                                                                                   \\
		\Psi_{\text{ASM}}\left(\Gamma(\textbf{x},\textbf{y})\right)      & = \widetilde{\mathcal{F}}\left[\Gamma(\textbf{u},\textbf{v})  \mathrm{H}\left(\lambda,\Delta x,M,d\right)\right]                                            \\
		\Psi_{\text{ASM}}^{-1}\left(\Gamma(\textbf{x},\textbf{y})\right) & = \widetilde{\mathcal{F}}\left[\Gamma(\textbf{u},\textbf{v}) \mathrm{H}\left(\lambda,\Delta x,M,-d\right)\right]                                            \\
		\mathrm{H}\left(\text{FMH}\right)                                & = \mathrm{H'}\left(\textbf{u},\textbf{v};\text{FMH}\right) \text{rect}\left(\frac{\textbf{u}}{2u_{\text{BL}}}\right) \text{rect}\left(\frac{\textbf{v}}{2v_{\text{BL}}}\right) \\
		\mathrm{H'}\left(\textbf{u},\textbf{v};\text{FMH}\right)                           & =
		\begin{cases}
			\exp^{\left(\mathrm{i}2\pi w(\textbf{u},\textbf{v})d\right)} & ,\text{if }u^{2} + v^{2} \leq \lambda^{2} \\
			0                                          & , \text{otherwise}
		\end{cases}                                                                                                            \\
		w\left( \textbf{u},\textbf{v} \right)                                              & = \left(\lambda^{-2} - \textbf{u}^{2} - \textbf{v}^{2} \right)^{1/2}                                                                                         \\
		(u_{\text{BL}},v_{\text{BL}})                                                    & = \frac{1}{\left[ \left( 2d(\Delta u,\Delta v)\right)^{2} + 1\right]^{1/2}\lambda} \\
	\end{aligned}
\end{equation}
\subsubsection{Unrolling of the GS algorithm}
Phase retrieval via iterative algorithms is slower as compared to neural networks. On the other hand black box neural networks are not explainable. In order to combine the benefits from both the worlds, we can unroll the iterative algorithm. Generally the unrolling is done to remove the iterative component of the iterative models and replace it with trainable neural network models. For the GS algorithm, the initial condition (random phase at the hologram plane) becomes a neural network and a loss function is introduced at the iterant position (Step 4 in \algoref[Algorithm 1]{algorithm:GS_Algorithm}) thereby training the phase retrieval network at the hologram plane\cite{Wu:21}. Further neural networks can be introduced to adjust the entire wavefront (amplitude and phase) at the hologram plane\cite{HosseinEybposh:20} and the SLM plane\cite{Peng2020}. Here we only include the phase retrieval neural network at the hologram plane for faster computations. Refer to \algoref[Algorithm 2]{algorithm:unrolled_GS_PINN} for the unrolling.

\subsubsection{Complex Valued Convolutional Neural Network}
Here we use a complex valued convolutional neural network \cite{Zhong2024} for approximating the initial phase at the hologram plane. We use a similar network as \cite{Zhong2024} due to fewer trainable parameters and comfortable computational burden for sensitivity analysis. \cite{Zhong2024} uses complex valued convolutional layers with skip connections. We modify the network to ensure that all input intensity image sizes and SLM pixel-resolutions within our hyperparameter bounds can be utilized for sensitivity analysis. Unlike \cite{Zhong2024} we only use the phase retrieval network at the hologram plane.

We use GS-PINN trained on both the forward models, GS algorithm as baseline for sensitivity analysis of forward model hyperparameters and comparisons between the forward models.

\subsection{Sensitivity Analysis of forward model hyperparameters}
\label{sec:method:sa_FMH}
To address the challenges inherent in designing new architectures, selecting optimal hardware configurations, and comparing neural network models with varying configurations, we propose leveraging Global Sensitivity Analysis (GSA) \cite{saltelli2008global}. GSA enables the examination of hyperparameters in forward models with respect to their impact on neural network performance. Neural networks are often regarded as ``black boxes", where understanding their internal operations is complex. However, efforts to gain insight into their behavior can foster a more systematic approach to neural network design. By applying GSA, we can evaluate parameter importance on a broader scale, contributing to a more transparent and explainable approach to neural network design. One promising avenue for understanding a neural network lies in analyzing how its performance responds to perturbations in inputs or parameters \cite{fel2021look,fel2023don}. Such an approach offers predictive value \cite{colin2022cannot}, enabling informed decisions in experimental and computational settings and allowing designers to prioritize parameters based on their influence on specific tasks. Sensitivity analysis (SA) methods can be particularly useful here, as they help elucidate the effects of changes in parameters on model outcomes.

SA techniques range from local, derivative-based methods to global, variance-based, stochastic approaches\cite{Zhang2022}. Local SA (LSA) techinques involves using partial derivatives of the outputs with respect to inputs in order to evaluate the impact of perturbation of parameters on its outputs. While effective for understanding sensitivity near a fixed point, LSA only explores behavior in small regions of uncertainty and usually considers changes to one or a few parameters at a time. This makes LSA inadequate for problems, where parameter interactions and non-linear input relationships are crucial. In contrast GSA varies all inputs simultaneously across their entire range capturing both individual and interaction effects on the outputs \cite{FernandezNavarro2017}. After the sampling of input parameters and the corresponding outputs, various metrics can be used to calculate the perturbation effects. Some metrics could be based on dependence measures like Csiszar f-divergences\cite{DaVeiga2014,csiszar1967information}, integral probability metrics\cite{Mller1997} or Hilbert-Schmidt independence criterion (HSIC)\cite{Novello2023,novello2022making}. Here we use variance measure based metric called Sobol’s indices due to its intuitive and straightforward interpretation \cite{fel2021look}. Below we explain Saltelli’s extension of Sobol's method for GSA based on Sobol/ ANOVA decomposition.
\begin{figure}
	\centering
        \includegraphics[height=9.3in]{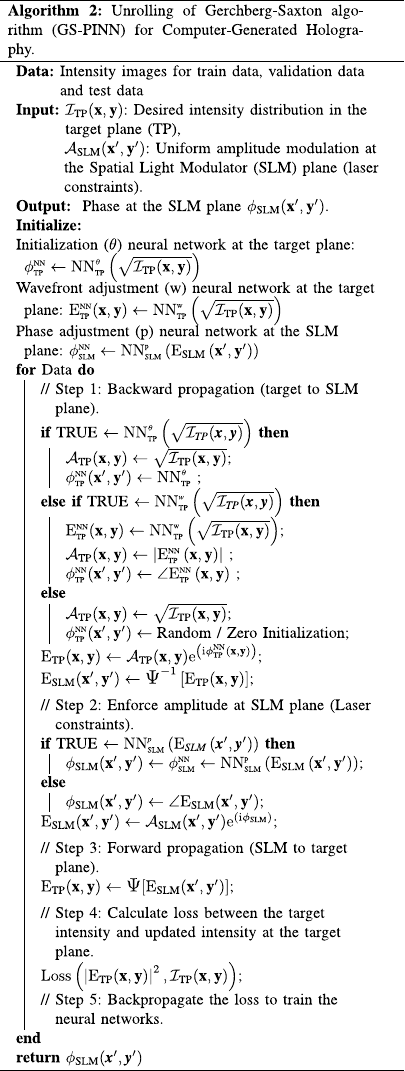}
	\caption*{}
	\label{algorithm:unrolled_GS_PINN}
\end{figure}

\balance
\subsubsection{Sobol's method} \cite{saltelli2008global} Given that a model is described by a function $Y = f(\mathrm{\textbf{X}})$, where $Y$ is a univariate model output and $\mathrm{\textbf{X}} = \{X_1, X_2, \dots, X_N\}$ are input parameters. We assume that $\mathrm{\textbf{X}}$ consists of $N$ independent and uniformly distributed variables within a unit hypercube, i.e. $X_i \in [0, 1]$ and $f(\mathrm{\textbf{X}})$ is an integrable function. With these assumptions, using ANOVA decomposition \cite{SALTELLI2010259}, $f(\mathrm{\textbf{X}})$ can be expressed as \cite{JECZMIONEK2022196}:
\begin{subequations}
	\begin{equation}
		\begin{aligned}
			 & Y = f_0 + \sum_{i} f_i(X_i) + \sum_{i < j} f_{i,j}(X_i, X_j) + \dots + f_{1,2,\dots,k},
		\end{aligned}
	\end{equation}
	Here integrant of all sum elements is zero.
	\begin{equation}
		\begin{aligned}
			 & \int_0^1 f_{i_1, i_2, \dots, i_s}(X_{i_1}, X_{i_2}, \dots, X_{i_s}) \, dX_{i_w} = 0.
		\end{aligned}
	\end{equation}
	where $1 \leq i_1<i_2<...<i_s \leq k$ and $i_w=\{i_1,i_2,...,i_s\}$. The functions $f_{i_1, i_2, \dots, i_s}$ are defined as:
	\begin{equation}
		\begin{aligned}
			f_0              & =  \mathbb{E}[Y],                                                                              \\
			f_i(X_i)         & =  \mathbb{E}_{\mathrm{\textbf{X}}_{\sim i}}[Y\mid X_i] - \mathbb{E}[Y],                       \\
			f_{i,j}(X_i,X_j) & =  \mathbb{E}_{\mathrm{\textbf{X}}_{\sim \{i, j\}}}[Y \mid X_i, X_j] - f_i(X_i) - f_j(X_j)     \\&
			- \mathbb{E}[Y]  &                                                                                            & ,
		\end{aligned}
	\end{equation}
	and similarly for higher-order terms. Here $\mathbb{E}_{\mathrm{\textbf{X}}_{i}}$ is the expected value over $X_i$ and $\mathbb{E}_{\mathrm{\textbf{X}}_{\sim i}}$ is the expected value over all except $X_i$. The relationship between the functions $f_{i_1, i_2, \dots, i_s}$ and partial variances is given by:
	\begin{equation}
		\begin{aligned}
			V_i     & = \mathbb{V}[f_i(X_i)]                                                                                \\
			        & = \mathbb{V}_{X_i} \big( \mathbb{E}_{\mathrm{\textbf{X}}_{\sim i}}[Y \mid X_i] \big),                 \\
			V_{i,j} & = \mathbb{V}[f_{i,j}(X_i, X_j)]                                                                       \\
			        & = \mathbb{V}_{X_i, X_j} \big( \mathbb{E}_{\mathrm{\textbf{X}}_{\sim \{i, j\}}}[Y \mid X_i, X_j] \big)
			- \mathbb{V}_{X_i} \big( \mathbb{E}_{\mathrm{\textbf{X}}_{\sim i}}[Y \mid X_i] \big)                            \\
			        & \quad - \mathbb{V}_{X_j} \big( \mathbb{E}_{\mathrm{\textbf{X}}_{\sim j}}[Y \mid X_j] \big),
		\end{aligned}
	\end{equation}
	The total variance $V(Y)$ is then expressed as:
	\begin{equation}
		V(Y) = \sum_i V_i + \sum_{i < j} V_{i,j} + \dots + V_{1,2,\dots,k}.
	\end{equation}
	Normalizing both sides of this equation by $V(Y)$, we obtain:
	\begin{equation}
		\sum_i S_i + \sum_{i < j} S_{ij} + \dots + S_{1,2,\dots,k} = 1,
	\end{equation}
	where $S_i$, $S_{ij}$, and higher-order terms represent normalized sensitivity indices.
	Here, the first-order sensitivity index:
	\begin{equation}
        \label{eq:Soblo_index_h_o}
		S_i = \frac{\mathbb{V}_{X_i} \big( \mathbb{E}_{\mathrm{\textbf{X}}_{\sim i}}[Y \mid \mathrm{\textbf{X}}_i] \big)}{V(Y)},
	\end{equation}
	quantifies the variance contribution of $X_i$, and the total-effect index:
	\begin{equation}
        \label{eq:Soblo_index_t_o}
		S_{T_{i}} = \frac{\mathbb{E}_{\mathrm{\textbf{X}}_{\sim i}} \big( \mathbb{V}_{X_i}[Y \mid \mathrm{\textbf{X}}_{\sim i}] \big)}{V(Y)} = 1 - \frac{\mathbb{V}_{\mathrm{\textbf{X}}_{\sim i}} \big( \mathbb{E}_{X_i}[Y \mid \mathrm{\textbf{X}}_{\sim i}] \big)}{V(Y)},
	\end{equation}
	quantifies the total effect (including first and higher order) of the factor $X_i$.
\end{subequations}
\subsubsection{First requirement}
To calculate the Sobol indices (\eqrefp{eq:Soblo_index_h_o} - \eqrefp{eq:Soblo_index_t_o}) some requirements need to be satisfied \cite{FernandezNavarro2017}.
\paragraph{Hyperparameter bounds}
The input variables should be contained within [0,1]. This requirement is generally satisfied because we can always use min-max normalization to satisfy the bounds. Our forward model hyperparameters include distance between SLM and Target plane ($d$), size of the SLM ($M$), pixel-pitch of the SLM ($\Delta x$), wavelength of light ($\lambda$) \figref{fig:fm_fmh}. For our experiments, we consider the SLM to have a square pixel layout with equal pixel-pitch and pixel-resolution in both dimentions. The hyperparameter bounds are as follows (units: meter (m)):\\
\begin{equation}
	\begin{aligned}
		\label{eq:FMH_bounds}
		\lambda  & =[\lambda_{\text{min}},\lambda_{\text{max}}]   & = & [200,1800](\textup{nm}) \\
		\Delta x & =[\Delta x_{\text{min}},\Delta x_{\text{max}}] & = & \normalfont{[4,80]}(\mu \textup{m})       \\
		M        & =[M_{\text{min}},M_{\text{max}}]               & = & [128,4000](\textup{pixels})                \\
		d        & =[d_{\text{min}},d_{\text{max}}]               & = & [0, 1.5](\textup{m})                  \\
	\end{aligned}
\end{equation}
The selection of hyperparameter bounds was made to explore the interactions between key parameters and assess their influence on model performance. The wavelength range of 200 nm - 1800 nm covers ultraviolet to near-infrared light, which aligns within the operational capabilities of most SLMs. The pixel-pitch range of 4 $\mu$m - 80 $\mu$m and pixel-resolution in the range of 128 to 4000 pixels is well within the manufacturing limits of commercial SLMs. The resulting first-order diffraction angles are confined between $0.14^{\circ}$ and $26.74^{\circ}$, as described by the grating equation $m\lambda=\Delta x \sin(\theta)$. Light propagation was modeled via band limited angular spectrum method upto a distance of 1.5 m. While these bounds provide a useful exploration of system behavior, further research is needed to assess the impact of extending these ranges or exploring alternative parameter settings. These parameter bounds were primarily chosen to investigate the systems behaviour under different scenarios, and the results obtained from these ranges reflect both the physical and computational limitations of the chosen configuration.

\subsubsection{Our approach}
\label{sec:methods_sa_fmh_approach}
\begin{figure}[H]
	\centering
	\includegraphics[width=2.5in]{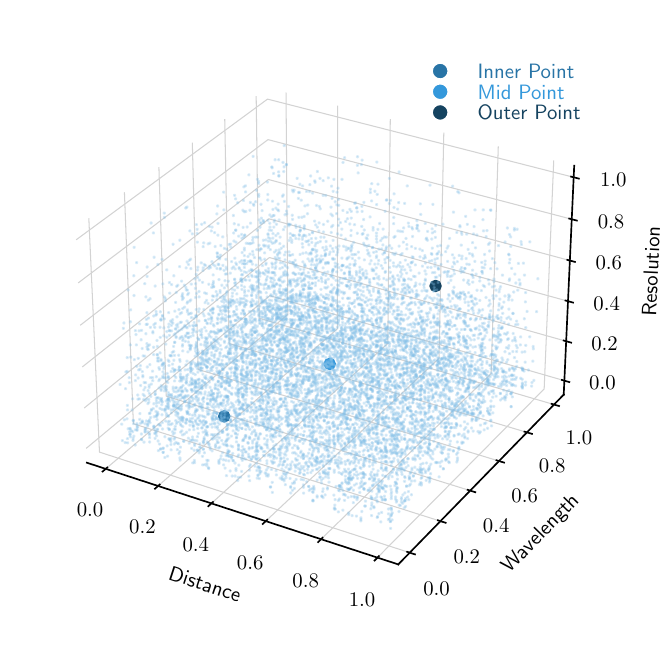}
	\caption{Normalized hyperparameter space with Inner, Mid and Outer points. Sampling for SA was performed using Saltelli's extension of Sobol's sequence \cite{sobol2001global,SALTELLI2002280,will_usher_2016_160164}. For $\textbf{\textup{h}}_{\textup{mid}}$, $N_{\textbf{\textup{h}}_{\textup{mid}}} = 1024$, generated $10240$ FMH configurations with $k = 4$ parameters. For
		$\textbf{\textup{h}}_{\textup{outer}}$ and $\textbf{\textup{h}}_{\textup{inner}}$, $N_{\textbf{\textup{h}}_{\textup{outer|inner}}} = 256$ producing 2560 experiments each. (Resolution: $M \Delta x$)}
	\label{fig:complete_hyperparameter_space}
\end{figure}
We now mathematically define the function for the unrolled Gerchberg-Saxton algorithm that we use for sensitivity analysis. From \algoref[Algorithm 2]{algorithm:unrolled_GS_PINN} we only use the initialization network $\left(\mathrm{NN}_{\scriptscriptstyle \text{TP}}^{\scriptscriptstyle \theta}\right)$. We train $\mathrm{NN}_{\scriptscriptstyle \text{TP}}^{\scriptscriptstyle \theta}$ on the inner, mid and outer points in the hyperparameter space (\figref{fig:complete_hyperparameter_space}, \tableref{table:inner_mid_outer_points}).
\\
\begin{table}[H]
	\caption{Inner, Mid and Outer points in the FMH space. $\langle ., . \rangle$ denotes the mean of the two values.}
	\label{table:inner_mid_outer_points}
	\begin{center}
		\begin{tabular}{|c|c|c|c|}
			\hline $\text{FMH}$ & $\mathbf{h}_{\text{mid}}$                                       & $\mathbf{h}_{\text{inner}}$                                    & $\mathbf{h}_{\text{outer}}$                                     \\
			\hline $\lambda$    & $\langle \lambda_{\text{min}}, \lambda_{\text{max}}\rangle$     & $\langle \lambda_{\text{min}}, \lambda_{\text{mid}} \rangle$   & $ \langle \lambda_{\text{mid}}, \lambda_{\text{max}} \rangle$   \\
			\hline $\Delta x$   & $ \langle \Delta x_{\text{min}}, \Delta x_{\text{max}} \rangle$ & $ \langle \Delta x_{\text{min}},\Delta x_{\text{mid}} \rangle$ & $ \langle \Delta x_{\text{mid}}, \Delta x_{\text{max}} \rangle$ \\
			\hline $M$          & $\langle M_{\text{min}}, M_{\text{max}} \rangle$                & $\langle M_{\text{min}}, M_{\text{mid}} \rangle$               & $\langle M_{\text{mid}}, M_{\text{max}} \rangle$                \\
			\hline $d$          & $\langle d_{\text{min}},d_{\text{max}} \rangle$                 & $\langle d_{\text{min}}, d_{\text{mid}}\rangle$                & $\langle d_{\text{mid}}, d_{\text{max}} \rangle$                \\
			\hline
		\end{tabular}
	\end{center}
\end{table}

From \algoref[Algorithm 3]{algorithm:finetuning_GS_SA_FMH}, the initial phase $\left(\phi_{\text{TP}}^{\scriptscriptstyle \text{NN}} \left( x, y \right)\right)$ is approximated by $\mathrm{NN}_{\scriptscriptstyle \text{TP}}^{\scriptscriptstyle \theta}$ at the target plane $\left(\text{TP}\right)$. The complex wavefront at the target plane is then given by $\mathrm{E}_{\text{TP}}(x, y)$ = $\mathrm{\mathcal{A}}_{\text{TP}}(x, y) \mathrm{exp}\left(\mathrm{i} \phi_{\text{TP}}^{\scriptscriptstyle \text{NN}} \left( x, y \right)\right)$. We propagate the complex field $\mathrm{E}_{\text{TP}}(x, y)$ from the target plane to the SLM using $\scalebox{1.1}{$\Psi$}^{-1}$.$ \text{ Here } \scalebox{1.1}{$\Psi$}^{-1}\left(\Gamma\left(x,y\right)\right)$ = $\widetilde{\mathcal{F}}\left[\mathcal{F}\left(\Gamma\left(x, y\right)\right)  \mathrm{H}\left(\lambda,\Delta x,M,d\right)\right]$. The complex wavefront at the SLM plane then becomes  $\mathrm{E}_{\text{SLM}}(x', y')$ = $\scalebox{1.1}{$\Psi$}^{-1}\left[\mathrm{E}_{\text{TP}}(x, y)\right]$. With uniform amplitude constraint at the SLM plane we extract the phase at SLM plane using $\phi_{\text{SLM}}$ = $\angle \mathrm{E}_{\text{SLM}}(x', y')$. Further, we propagate the updated SLM plane complex wavefront $\mathrm{E}_{\text{SLM}}(x', y')$ = $\mathrm{exp}\left(\mathrm{i}\phi_{\text{SLM}} \right)$ back to the target plane using $\scalebox{1.1}{$\Psi$}\left(\Gamma(x',y')\right)$ = $\widetilde{\mathcal{F}}\left[\mathcal{F}\left(\Gamma(x', y')\right)  \mathrm{H}\left(\lambda,\Delta x,M,-d\right)\right]$ to get the updated target complex wavefront $\mathrm{E}_{\text{TP}}(x, y)$= $\scalebox{1.1}{$\Psi$}[\mathrm{E}_{\text{SLM}}(x', y')]$. We use the procedure outlined in \algoref[Algorithm 3]{algorithm:finetuning_GS_SA_FMH} to finetune the network $\mathrm{NN}_{\scriptscriptstyle \text{TP}}^{\scriptscriptstyle \theta}$ on different hyperparameter configurations. Our Accuracy functions can then be defined as:
\begin{subequations}
	\begin{equation}
		\begin{aligned}
			 & \overline{\mathrm{PSNR}}\left(\left|\mathrm{E}_{\text{TP}}(x, y)\right|^{2},\mathrm{\mathcal{I}}_{\text{TP}}(x, y)\right)_{\textup{Test}},    \\
			 & \overline{\mathrm{SSIM}}\left(\left|\mathrm{E}_{\textup{TP}}(x, y)\right|^{2},\mathrm{\mathcal{I}}_{\textup{TP}}(x, y)\right)_{\textup{Test}} \\
			\label{eq:acuracy_functions}
		\end{aligned}
	\end{equation}
    where,
	\begin{equation}
		\begin{aligned}
			\mathrm{E}_{\text{TP}}(x, y)                                                & =
			\scalebox{1.2}{$\Psi$}_{\textrm{ASM}}[\mathrm{\mathcal{A}}_{\text{SLM}}(x', y') \mathrm{exp}{\left(\mathrm{i}\phi_{\text{SLM}} \right)}]                                                                                 \\
			\phi_{\text{SLM}}                                                           & =
			\angle \scalebox{1.2}{$\Psi$}^{-1}_{\text{ASM}}
			\left[\sqrt{\mathrm{\mathcal{I}}_{\text{TP}}(x, y)}\mathrm{exp}{\left(\mathrm{i} \phi_{\text{TP}}^{\scriptscriptstyle \text{NN}} \left( x, y \right)^{\mathcal{E}: 1 \rightarrow 5}\right)}\right]                                              \\
			\phi_{\text{TP}}^{\scriptscriptstyle \text{NN}} \left( x, y \right)^{\mathcal{E}: 1 \rightarrow 5} & = \mathrm{NN}_{\scriptscriptstyle \text{TP}}^{\scriptscriptstyle \theta}\left(\sqrt{\mathrm{\mathcal{I}}_{\text{TP}}(x, y)}\right)^{\mathcal{E}: 1 \rightarrow 5} \\
			\label{eq:acuracy_functions_explain}
		\end{aligned}
	\end{equation}
\end{subequations}
where $\mathrm{NN}_{\scriptscriptstyle \text{TP}}^{\scriptscriptstyle \theta}\left(\sqrt{\mathrm{\mathcal{I}}_{\text{TP}}(x, y)}\right)^{\mathcal{E}: 1 \rightarrow 5}$ is the initialization network optimized for 5 epochs ($\mathcal{E}: 1 \rightarrow 5$). Optimization was done upto 5 epochs due to computational constraints.  Peak Signal to Noise Ratio (PSNR) and Structural Similarity Index Measure (SSIM) \cite{SSIM} have the usual definitions. We use these functions to calculate the Sobol indices.
\begin{figure}[!t]
	\centering
        \includegraphics[height=9.0in]{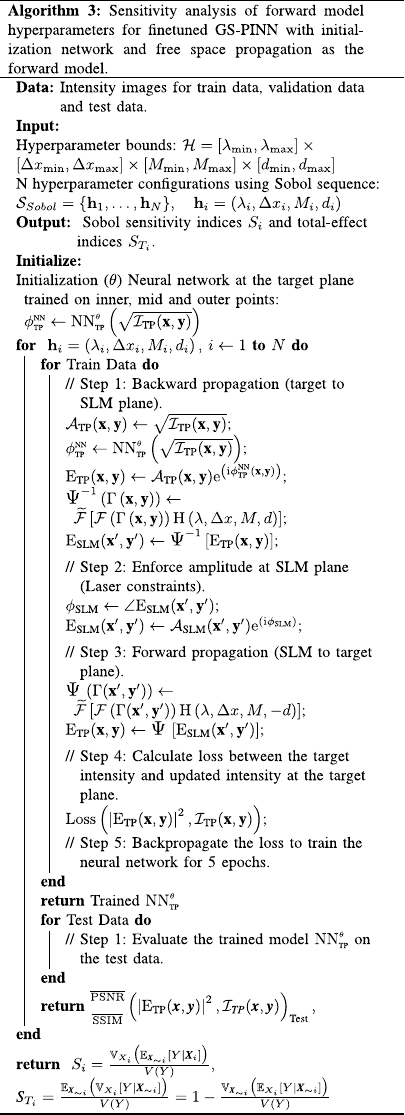}
	\caption*{}
	\label{algorithm:finetuning_GS_SA_FMH}
\end{figure}
\subsubsection{Second requirement}
Here we discuss the second condition in order to calculate Sobol's indices.
\paragraph{Square integrability}
Refer to \algoref[Algorithm 3]{algorithm:finetuning_GS_SA_FMH} and \secref{sec:methods_sa_fmh_approach}. The accuracy function \eqrefp{eq:acuracy_functions} - \eqrefp{eq:acuracy_functions_explain} must be square integrable. Here the output of the initialization neural network $\mathrm{NN}_{\scriptscriptstyle \text{TP}}^{\scriptscriptstyle \theta}$ is a phase mask which is bounded in $[-\pi,\pi]$. $ \scalebox{1.2}{$ \Psi$} \textup{ and } \scalebox{1.2}{$\Psi$}^{-1}$ are energy preserving transformations. SSIM is bounded in $[0,1]$. To prevent the mean squared error (MSE) in the denominator of the PSNR calculation from reaching zero, a small epsilon ($\epsilon$) value is added, where $\epsilon$ is an arbitrarily small positive value. With this in mind, we ensure that the accuracy functions used in calculations of the Sobol indices are finite and square-integrable.

\subsection{Sensitivity analysis for different forward models}
\label{sec:method:sa_FM}
The performance of neural networks in phase retrieval tasks is sensitive to the choice of forward model. Each FM has unique characteristics and hardware-specific configurations, including wavelength, SLM pixel-resolution, pixel-pitch, and propagation distance. This sensitivity creates a significant challenge: neural networks trained on one FM often fail to generalize when applied to other configurations. Experimentalists rely on FMs to simulate physical systems, but selecting an inappropriate FM can lead to suboptimal performance or poor generalization to real-world setups. Addressing this challenge requires a systematic evaluation of FMs to uncover their effects on learning and generalization.
\subsubsection{Our approach}
\begin{figure}[!t]
	\centering
	\includegraphics[width=3.5in]{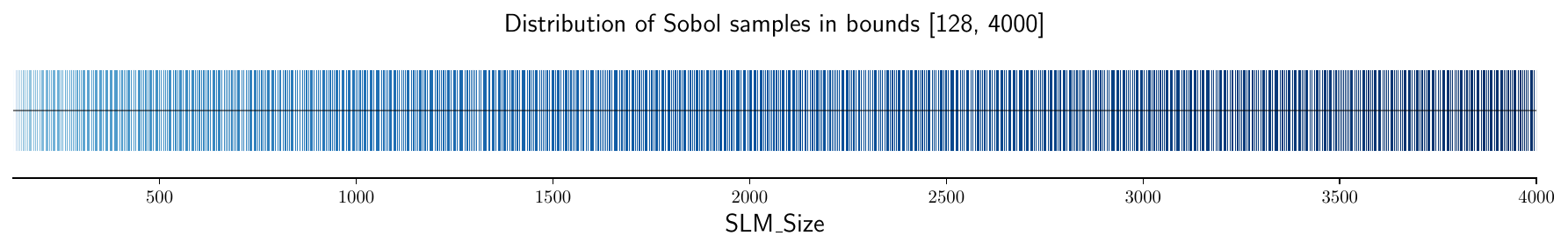}
	\caption{Sobol's samples for SLM pixel-resolution within the bounds [128,4000] to analyze sensitivity of FMs.}
	\label{fig_4}
\end{figure}
To address this, we systematically evaluate and compare FMs such as Fourier holography and free space propagation. Fourier holography and free space propagation FMs have a common hyperparameter, namely the size of the SLM. Here we ask ourselves, how do different FMs affect the performance of GS-PINN given the size of SLM? (\algoref[Algorithm 4]{algorithm:finetuning_GS_PINN_FM}). Here we train the neural network with two different forward models having same parameters as $\textbf{h}_{\text{mid}}$. We call these base models trained with different forward models $\texttt{base\_fourier}$ and $\texttt{base\_free}$. While training we fix the neural network hyperparameters and FMH for both the base models. We use quasi-random Sobol's sequence \cite{will_usher_2016_160164} to sample 1024 points within the SLM pixel-resolution bounds [128,4000] (\figref{fig_4}). We use these new SLM hyperparameters to fine-tune the $\texttt{base\_fourier}$ and $\texttt{base\_free}$ models for 5 epochs. We call these models $\texttt{base\_fourier\_fourier}$ and $\texttt{base\_free\_free}$. We also interchange the forward model and fine-tune the base models for 5 epochs with the same SLM hyperparameters. We call these models $\texttt{base\_fourier\_free}$ and $\texttt{base\_free\_fourier}$ respectively (\algoref[Algorithm 4]{algorithm:finetuning_GS_PINN_FM}). We perform similar experiments using the Gerchberg-Saxton algorithm for both the forward models.

This analysis serves as a guide for selecting optimal FMs. It also provides critical insights into how FMs influence neural network training and deployment. These findings support decision-making while choosing neural networks for CGH in practical applications.

\begin{figure}[!t]
	\centering
        \includegraphics[height=9in]{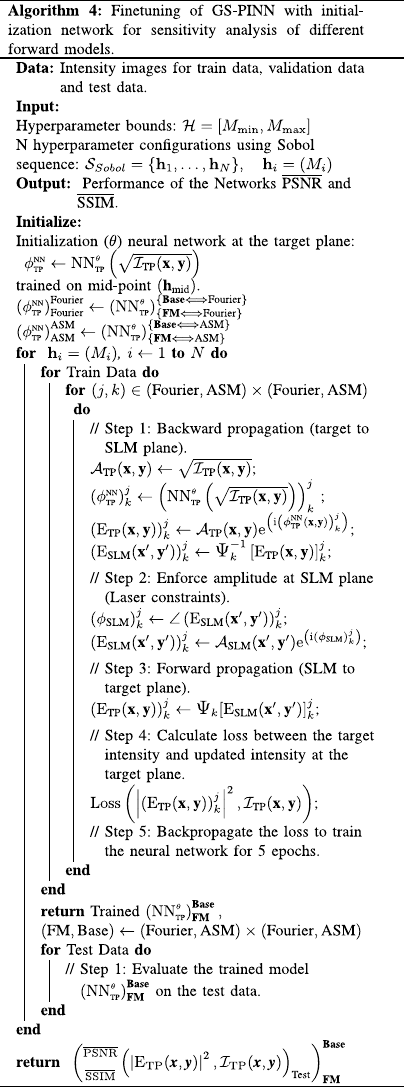}
	\caption*{}
	\label{algorithm:finetuning_GS_PINN_FM}
\end{figure}

\subsection{Metric}
\label{sec:method:metric}
The design and optimization of neural networks are often conducted with fixed forward model hyperparameters (FMH), limiting the ability to evaluate network performance across varying FMH configurations. Networks trained on one FMH may not generalize well to others, complicating comparisons between networks and their robustness to FMH variations (\secref{sec:results:composite_metric}). This highlights the need for fair and consistent evaluation metrics that account for these differences.

To address these challenges, we propose three distinct metrics designed to provide fair and context-sensitive evaluations of neural network performance across varying FMH configurations. The GS-Weighted Metric focuses on comparisons within the same FMH, leveraging a standard algorithm and emphasizing the performance of networks relative to the FMH on which they were designed. The Generalization Metric extends this scope by evaluating networks on independent FMH, enabling comparisons that are decoupled from both the specific FMH used during design and any standard algorithm. Finally, the Resilience Metric shifts the focus to local hyperparameter variations, assessing the network's robustness within its immediate hyperparameter space without involving direct comparisons between different networks. These metrics collectively address the need for fair evaluation tailored to diverse experimental requirements.

\subsubsection{GS-Weighted metric}
The GS algorithm provides a standardized baseline for evaluating neural network performance. However, the complexity induced by FMH in the GS algorithm and neural networks is uncorrelated (\secref{sec:results:composite_metric}). We assume this is due to the ill-posed nature of the inverse problem. This further induces difficulty in comparing different neural networks (NN) with GS algorithm as the baseline.

For example, a network trained on one FMH may perform poorly on another with higher complexity, potentially biasing the results. Despite this limitation, GS-based comparisons are beneficial for understanding network adaptability to varying FMH and providing a reference point relative to a standard iterative algorithm. While imperfect, this approach offers valuable insights into network performance consistency and adaptability.
\begin{subequations}
	\begin{equation}
		\text{GS-Weighted Metric} (\Upsilon_{\textup{gsw}}) = \frac{1}{N} \sum_{i=1}^{N} \frac{P_i}{\text{GS}_i},\\
	\end{equation}
	where:
	\begin{align*}
		N             & = \textup{Number of FMH configurations}                   \\
		\text{GS}_{i} & = \textup{Normalized GS algorithm performance on } i^{th} \textup{ FMH} \\
		P_{i}         & = \textup{Normalized NN performance on } i^{th} \textup{ FMH.}
	\end{align*}

	\subsubsection{Generalization metric}
	The generalization metric addresses the limitations of GS algorithm comparisons by evaluating networks on fixed, standardized FMH configurations (inner, center, and outer points) that are independent of the training FMH and GS algorithm. By keeping neural network hyperparameters constant and retraining networks on these FMH, the metric eliminates GS dependence and measures generalization capacity across independent FMHs. This approach ensures networks are tested on FMHs they were not optimized for, providing a fair assessment of their adaptability to new configurations. We use the FMH associated to $\textbf{\textup{h}}_{\textup{inner}}$, $\textbf{\textup{h}}_{\textup{mid}}$ and $\textbf{\textup{h}}_{\textup{outer}}$ (\tableref{table:inner_mid_outer_points}, \eqrefp{eq:FMH_bounds}).
	\begin{equation}
		\text{Generalization Metric} (\Upsilon_{\textup{gm}}) = \frac{1}{3}  \sum_{j} P_\textup{j}
	\end{equation}
	where $P_{j} (j \in \{\textbf{\textup{h}}_{\textup{inner}}, \textbf{\textup{h}}_{\textup{mid}}, \textbf{\textup{h}}_{\textup{outer}}\})$ is the normalized NN performance on the $j^{th}$ FMH.

	However, this metric has its limitations. The relative complexity of the fixed FMHs may still vary between networks, potentially favoring one network over another. For instance, the complexity of the standardized FMHs could be inherently easier for one network to handle than another, introducing a bias. Despite this, the generalization metric effectively captures a network's ability to perform across a representative range of FMH configurations, offering valuable insights into its robustness and adaptability while overcoming the dependency issues inherent in GS algorithm-based comparisons.

	\subsubsection{Resilience metric}
	The resilience metric evaluates a network's robustness to local perturbations in its FMH by assessing its performance within a small neighborhood around a specific FMH configuration ($\textup{FMH}_{1}$) that the network was trained on. To compute this, we sample a set of perturbed FMH using a Sobol sequence, $\Delta \textup{FMH} \sim \mathcal{S}(\textup{FMH}_{1}, \sigma)$ where the perturbed FMH configuration is constrained within a neighborhood defined by $ \pm \sigma$ around the original FMH configuration $\textup{FMH}_{1}$. Each perturbed FMH is defined as:
	\begin{equation}
		\textup{FMH}_{i} = \Delta \textup{FMH}_{i}
	\end{equation}
	where $\Delta \textup{FMH}_{i}$ is drawn from the Sobol sequence, ensuring that each perturbed FMH is within the fixed neighborhood $\pm \sigma$ around $\textup{FMH}_{1}$. The network’s performance on these perturbed FMHs is quantified using the normalized accuracy/performance scores $P(\textup{FMH}_{i})$. The resilience metric is then given by:
	\begin{equation}
		\textup{Resilience Metric}(\Upsilon_{\textup{r}}) = 1 - \frac{1}{N} \sum_{i=1}^{N} \frac{\left( P(\textup{FMH}_{i}) - P(\textup{FMH}_{1})\right)^{2}}{P(\textup{FMH}_{1})}
	\end{equation}
	where $N$ is the total number of sampled FMHs. The metric measures the variability in performance accross the neighborhood of FMHs relative to the performance on the reference $\textup{FMH}_{1}$. A value of $\Upsilon_{\textup{r}}=1$ indicates perfect resilience, where the networks performance is stable across local variations, while lower values suggest greater sensitivity to perturbations.
	Unlike the GS algorithm and generalization metrics, this approach is independent of direct comparisons between networks, focusing solely on each network's performance within its immediate hyperparameter surroundings. This provides valuable insights into a network's perturbation invariance, which can be critical for applications requiring stability under local variations \cite{chen2022fourier,huang2023self}.  While the resilience metric does not facilitate comparisons across networks, limiting its use for broader benchmarking, it addresses the limitations of the GS and generalization metrics by offering a localized evaluation.

	\subsubsection{Composition of the metric}
	Depending on the requirements of the CGH experiments the combined composite metric can be weighted.
	\begin{equation}
		\Upsilon = \alpha (\Upsilon_{\textup{gsw}}) + \beta (\Upsilon_{\textup{gm}}) + \gamma (\Upsilon_{\textup{r}})
	\end{equation}
\end{subequations}
where $\alpha, \beta, \gamma$ are the weights for different components of the composite metric.
\subsubsection{Limitations}
Directly comparing neural networks or benchmarking them against the GS algorithm is unreliable. Refer to \secref{sec:results:composite_metric} and \secref{sec:discussion:composite_metric}. Attempting to model these dependencies using neural networks or copulas \cite{nelsen2006introduction} introduces additional challenges, such as reduced interpretability and less clarity in benchmarking claims. This trade-off highlights the difficulty of designing evaluation metrics that are both fair and interpretable. The composite metric integrates GS-weighted, generalization, and resilience metrics to ensure fairness, adaptability, and robustness in network evaluation. While it reduces unreliable comparisons and speculative claims, it does not fully resolve biases arising from the independent FMH-dependent complexities of the GS algorithm and neural networks. Despite these challenges, the composite metric serves as a practical and well-rounded compromise.
\begin{figure}[H]
	\centering
	\includegraphics[width=3.5in]{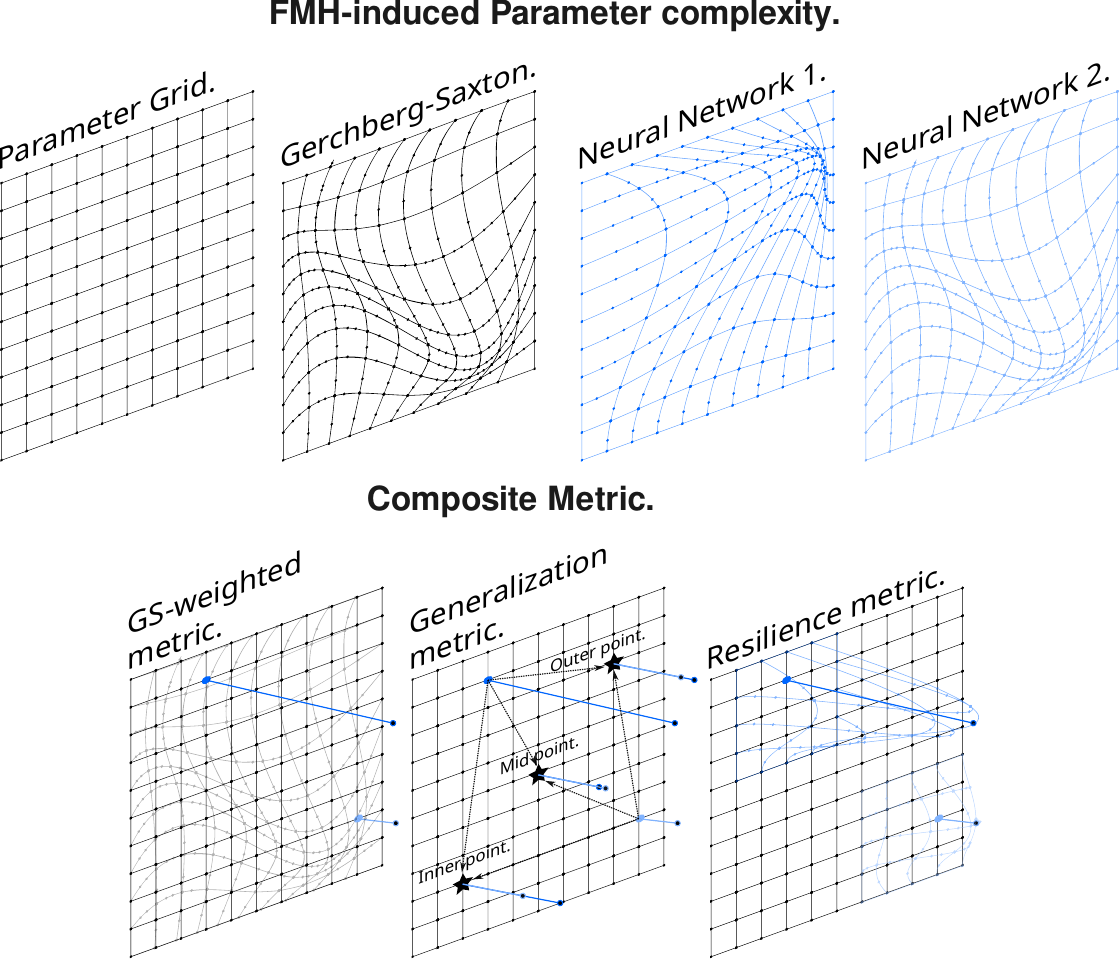}
	\caption{FMH-induced parameter complexity and the composite metric. Top panel represents the inconsistency in performance of different algorithms for similar set of FMH configurations, complicating benchmarking. In the bottom panel, the composite metric addresses variability in model performance across similar FMH configurations, enabling more reliable benchmarking.}
	\label{fig:metric}
\end{figure}

\section{Results}
\label{sec:results}
\subsection{Sensitivity Analysis of Forward model hyperparameters (FMH) for ASM:}
\label{sec:results:sa_fmh}
To evaluate the effect of FMH on the performance of GS-PINN and GS algorithm, GS-PINN was trained with fixed $\textbf{\textup{h}}_{\textup{mid}}$, $\textbf{\textup{h}}_{\textup{outer}}$, $\textbf{\textup{h}}_{\textup{inner}}$ FMH values as described in the \algoref[Algorithm 3]{algorithm:finetuning_GS_SA_FMH}. For $\textbf{\textup{h}}_{\textup{mid}}$, we sampled $N_{\textbf{\textup{h}}_{\textup{mid}}} = 1024$, generating $N_{\textbf{\textup{h}}_{\textup{mid}}}(2k + 2) = 10240$ experiments with $k = 4$ parameters for sensitivity analysis (\figref{fig:complete_hyperparameter_space}). Similarly, for
$\textbf{\textup{h}}_{\textup{outer}}$ and $\textbf{\textup{h}}_{\textup{inner}}$ we sampled $N_{\textbf{\textup{h}}_{\textup{outer|inner}}} = 256$ producing 2560 experiments each to calculate sensitivity indices and access stability. Sampling was done using Saltelli's extension of the Sobol's sequence \cite{sobol2001global,SALTELLI2002280,will_usher_2016_160164}. The models were trained using the DIV2K dataset (High Resolution) \cite{agustsson2017ntire}, with 800 images from the DIV2K\_train\_HR subset used for training and validation, split in a ratio of 87.5\% for training and 12.5\% for validation. The testing set consisted of 100 images from the DIV2K\_test\_HR subset. Training was performed using the Adam optimizer with a learning rate of 0.001 and a weight decay of 0.001, while the loss function employed was Mean Squared Error (MSE) for simplicity. The base models were trained for a total of 500 epochs, and the epoch corresponding to the highest validation score was selected for further experiments. The neural network hyperparameters remained consistent across all experiments. The neural network models were trained and evaluated on a high-performance computing (HPC) cluster based on HTCondor, equipped with a diverse set of NVIDIA GPUs. Utilizing the GPU cluster enabled efficient parallel execution of the large-scale simulations required for Sobol’s sensitivity analysis, significantly improving computational efficiency and scalability. 
\subsubsection{GS-PINN}
\begin{figure}[!ht]
	\centering
	\includegraphics[width=3.5in]{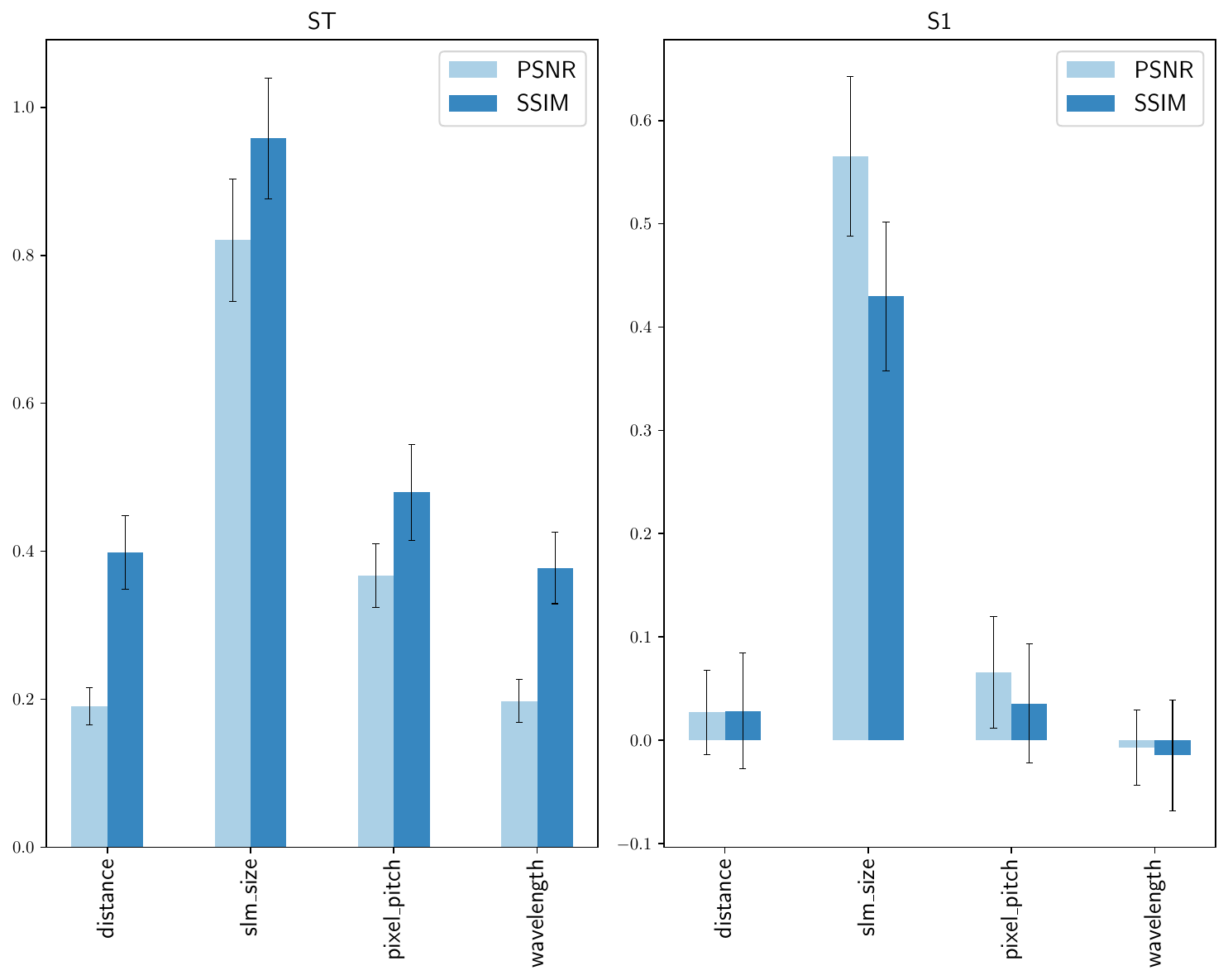}
	\caption{Sensitivity analysis (SA) for GS-PINN at $\textbf{\textup{h}}_{\textup{mid}}$ (10240 FMH configurations). The bar charts display the Sobol indices for PSNR (light blue) and SSIM (dark blue). For SA the accuracy functions \eqrefp{eq:acuracy_functions}, \eqrefp{eq:acuracy_functions_explain} were evaluated and averaged over 100 test images for each FMH configuration. The left panel (ST) shows the total-order indices reflecting the overall contribution of parameters. The right panel (S1) highlights the first-order indices representing the direct contribution of inputs. Small negative indices can be treated as zero. Error bars (95\% confidence) indicate uncertainty in the sensitivity indices.}
	\label{GSPINN_SA_50}
\end{figure}

\begin{figure}[H]
	\centering
	\includegraphics[width=3.5in]{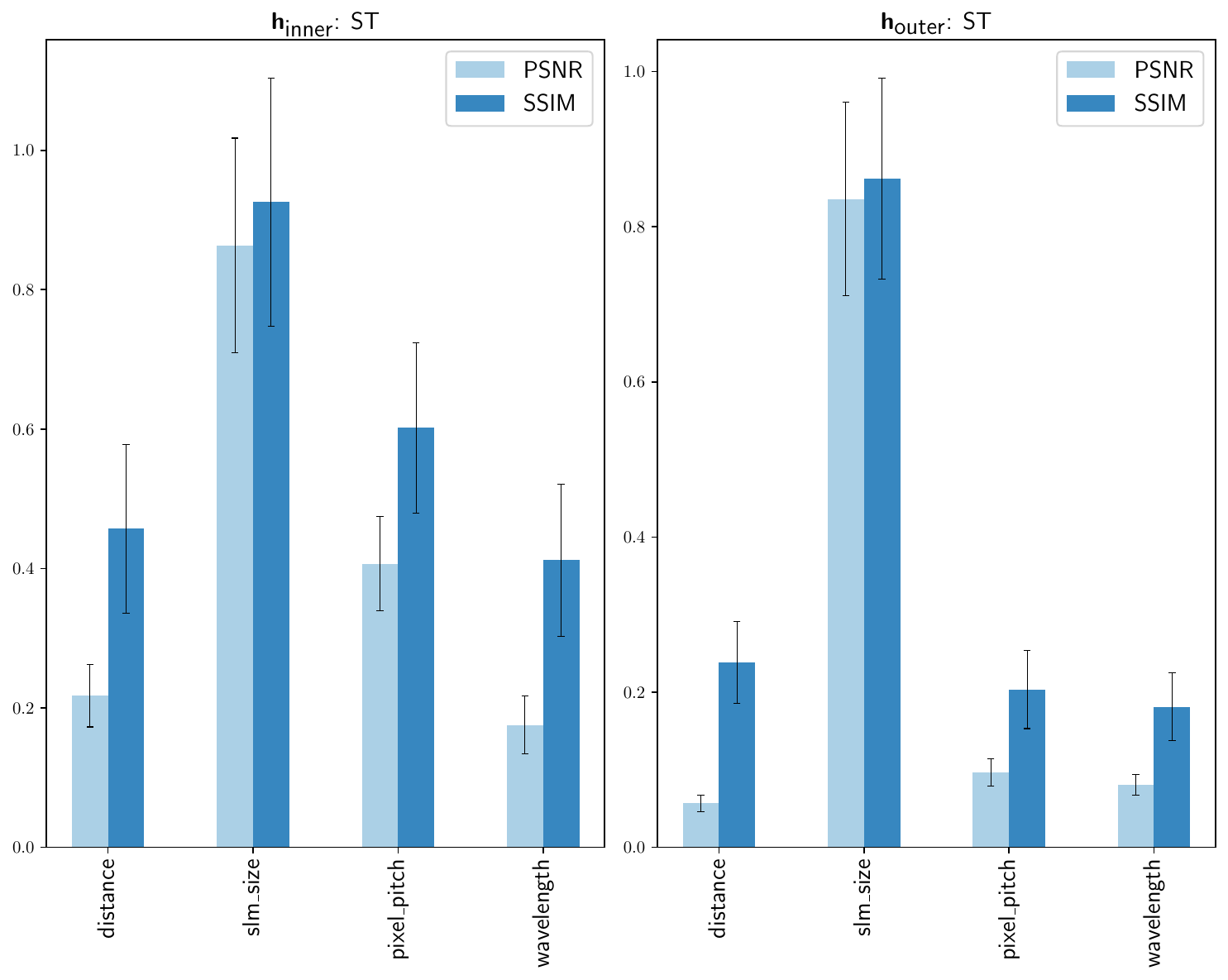}
	\caption{Sensitivity analysis (SA) for GS-PINN at $\textbf{\textup{h}}_{\textup{inner}}$ and $\textbf{\textup{h}}_{\textup{outer}}$ (2560 FMH configurations each). The bar charts display the total-order Sobol indices reflecting the overall contribution of parameters for PSNR (light blue) and SSIM (dark blue) accuracy functions \eqrefp{eq:acuracy_functions}. For SA the accuracy functions were evaluated and averaged over 100 test images for each FMH configuration. Error bars (95\% confidence) indicate uncertainty in the sensitivity indices.}
	\label{GSPINN_SA_25_t1}
\end{figure}

\paragraph{$\textbf{\textup{h}}_{\textup{mid}}$}
Total (ST), first-order (S1) and second-order (S2) were computed. The first-order Sobol index measures an input's direct contribution to output variance, the second-order index quantifies the contribution of interactions between two inputs, and the total-order index accounts for an input's overall contribution, including all interactions. For absolute values refer \tableref{tab:GSPINN_sa_50_PSNR}, \tableref{tab:GSPINN_sa_50_SSIM}. The parameter SLM pixel-resolution exhibited the highest sensitivity, contributing the most to variance in neural network performance. This was followed by SLM pixel-pitch, propagation distance and wavelength of light (\figref{GSPINN_SA_50}). Due to limited parameter samples, S2 indices show instability as reflected in wide confidence intervals (\tableref{tab:GSPINN_sa_50_PSNR}, \tableref{tab:GSPINN_sa_50_SSIM}).
\paragraph{$\textbf{\textup{h}}_{\textup{inner|outer}}$}
Similar to $\textbf{\textup{h}}_{\textup{mid}}$, SLM pixel-resolution remained the most influential parameter accross ${\textbf{\textup{h}}_{\textup{outer}}}$
and ${\textbf{\textup{h}}_{\textup{inner}}}$, with a consistent relative sensitivity profile for other FMH parameters (\figref{GSPINN_SA_25_t1}). However, absolute ST values for propagation distance, pixel-pitch and wavelength decreased for ${\textbf{\textup{h}}_{\textup{outer}}}$ as compared to ${\textbf{\textup{h}}_{\textup{inner}}}$. The S1 and S2 indices exhibit instability, evident from their broad confidence intervals, due to the limited number of parameter samples. For absolute values refer to \tableref{tab:GSPINN_sa_25_PSNR}, \tableref{tab:GSPINN_sa_25_SSIM}, \tableref{tab:GSPINN_sa_75_PSNR}, \tableref{tab:GSPINN_sa_75_SSIM}.
\paragraph{Accuracy function sensitivity}
PSNR and SSIM followed similar relative sensitivity profiles for $\textbf{\textup{h}}_{\textup{mid}}$ and ${\textbf{\textup{h}}_{\textup{inner}}}$. For ${\textbf{\textup{h}}_{\textup{outer}}}$, the absolute sensitivity rankings (\tableref{tab:GSPINN_sa_75_PSNR}, \tableref{tab:GSPINN_sa_75_SSIM}) for propagation distance and pixel-pitch were interchanged between PSNR and SSIM. However, their confidence intervals overlap, reflecting their similar contributions (\figref{GSPINN_SA_50}, \figref{GSPINN_SA_25_t1}).
\paragraph{Interaction effects}
The sum of ST indices of all the parameters is greater than 1, reflecting higher order interactions. The S1 index for SLM pixel-resolution is the highest followed by the SLM pixel-pitch. The S2 index for SLM pixel-resolution and pixel-pitch are also higher as compared to other interaction parameters. Interaction term between wavelength of light and propagation distance cause the least variance in the performance of GS-PINN as compared to other parameters. However, it is important to note that the confidence intervals for S2 indices indicate a high degree of uncertainty in these estimates. As such, the S2 analysis should be treated as preliminary and interpreted with caution. Further refinement of the analysis is necessary for more robust conclusions (\tableref{tab:GSPINN_sa_50_PSNR}, \tableref{tab:GSPINN_sa_50_SSIM}, \figref{GSPINN_SA_50}, \figref{GSPINN_SA_25_t1}).


\subsubsection{GS-Algorithm}
\begin{figure*}[!ht]
	\centering
	\includegraphics[width=5.5in]{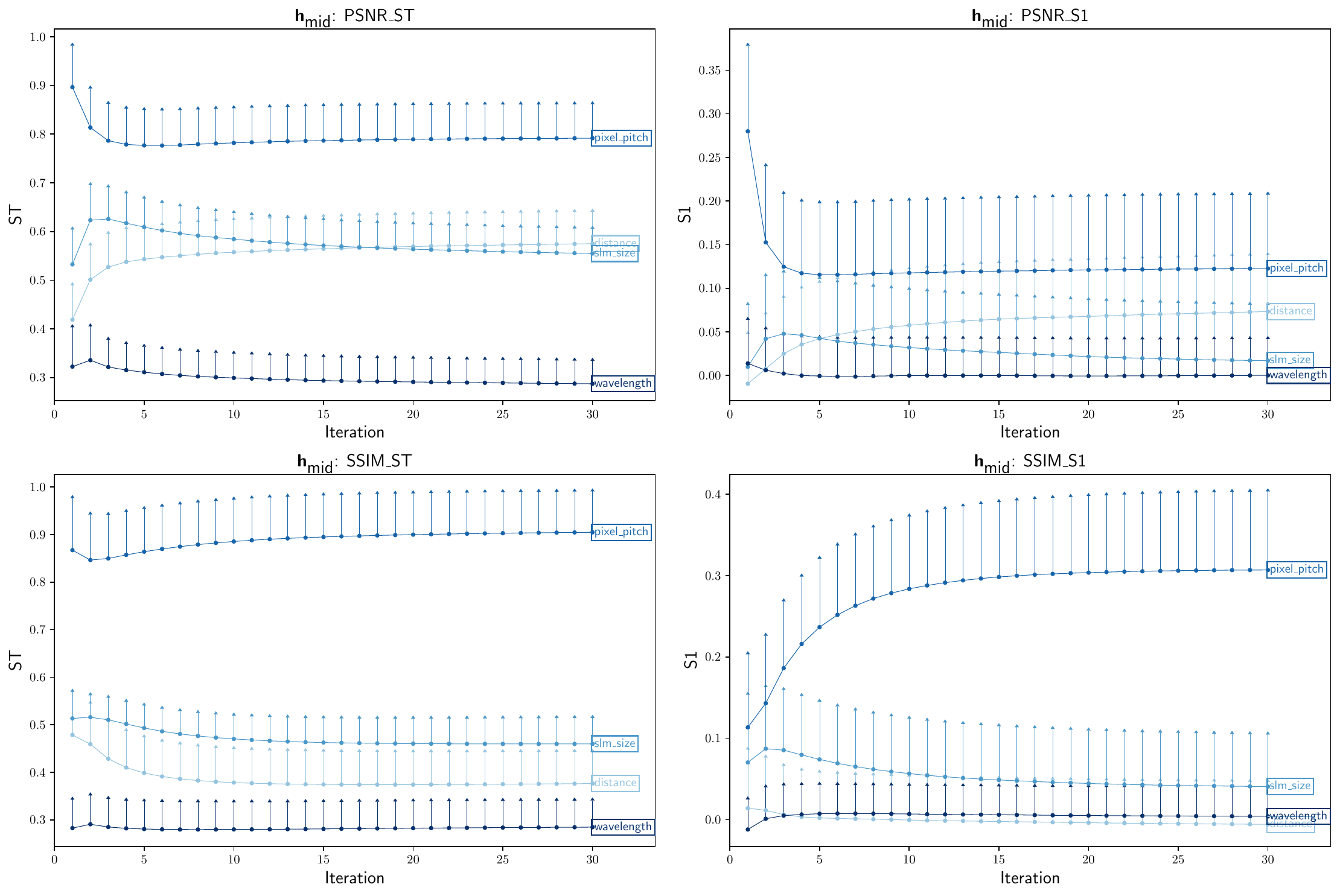}
	\caption{Sensitivity Analysis (SA) for $\textbf{\textup{h}}_{\textup{mid}}$ (10240 FMH configurations) over all iterations for PSNR and SSIM accuracy functions. For SA the accuracy functions were evaluated and averaged over 100 test images for each FMH configuration at each iteration. The left hand panel shows total-order (ST) indices reflecting the overall parameter contributions. The right hand panel displays first-order (S1) indices indicating the direct contributions of parameters. The top and bottom panels correspond to PSNR and SSIM accuracy functions respectively, as defined in \eqrefp{eq:acuracy_functions}, \eqrefp{eq:acuracy_functions_explain}. Error bars represent the upper limit of the 95\% confidence intervals for clarity.}
	\label{fig:0001_50_gs_SA_iterations_all}
\end{figure*}

We employed the same FMH samples utilized in the GS-PINN experiments to evaluate the performance of the GS algorithm. The GS algorithm was executed for up to 30 iterations across the FMH configurations associated with ${\textbf{\textup{h}}_{\textup{mid}}}$ (10240 samples), ${\textbf{\textup{h}}_{\textup{inner}}}$ (2560 samples) and ${\textbf{\textup{h}}_{\textup{outer}}}$ (2560 samples). Instead of using a constant or quadratic initial phase, a random initial phase was adopted to mimic the output of GS-PINN. To ensure the replicability of the sensitivity analysis, the random number generator was seeded. We investigated how the sensitivity indices evolve with the iterations of the GS algorithm and how they differ with those observed in the GS-PINN framework. Refer to \algoref[Algorithm 5.]{algorithm:finetuning_GS_FMH}
\paragraph{$\textbf{\textup{h}}_{\textup{mid}}$}
The total-order (ST), first-order (S1), and second-order (S2) sensitivity indices were calculated to evaluate the influence of input parameters on the performance of the GS algorithm. Due to the limited number of experimental trials, the S2 indices exhibited high instability and are thus excluded from further analysis. Consequently, we focus on the S1 and ST indices for interpretation (\figref{fig:0001_50_gs_SA_iterations_all}, \tableref{tab:GS_sa_50_PSNR}, \tableref{tab:GS_sa_50_SSIM}).

For both accuracy functions analyzed, the ST indices revealed a consistent ranking of parameters in terms of their contribution to variance. Among the parameters, the pixel pitch was identified as the most significant contributor to performance variability, whereas wavelength exhibited the lowest contribution. Notably, the contributions of propagation distance and SLM pixel-resolution showed overlapping confidence intervals, indicating similar levels of influence on the observed variance in GS algorithm performance.

\begin{figure*}[!ht]
	\centering
	\includegraphics[width=5.5in]{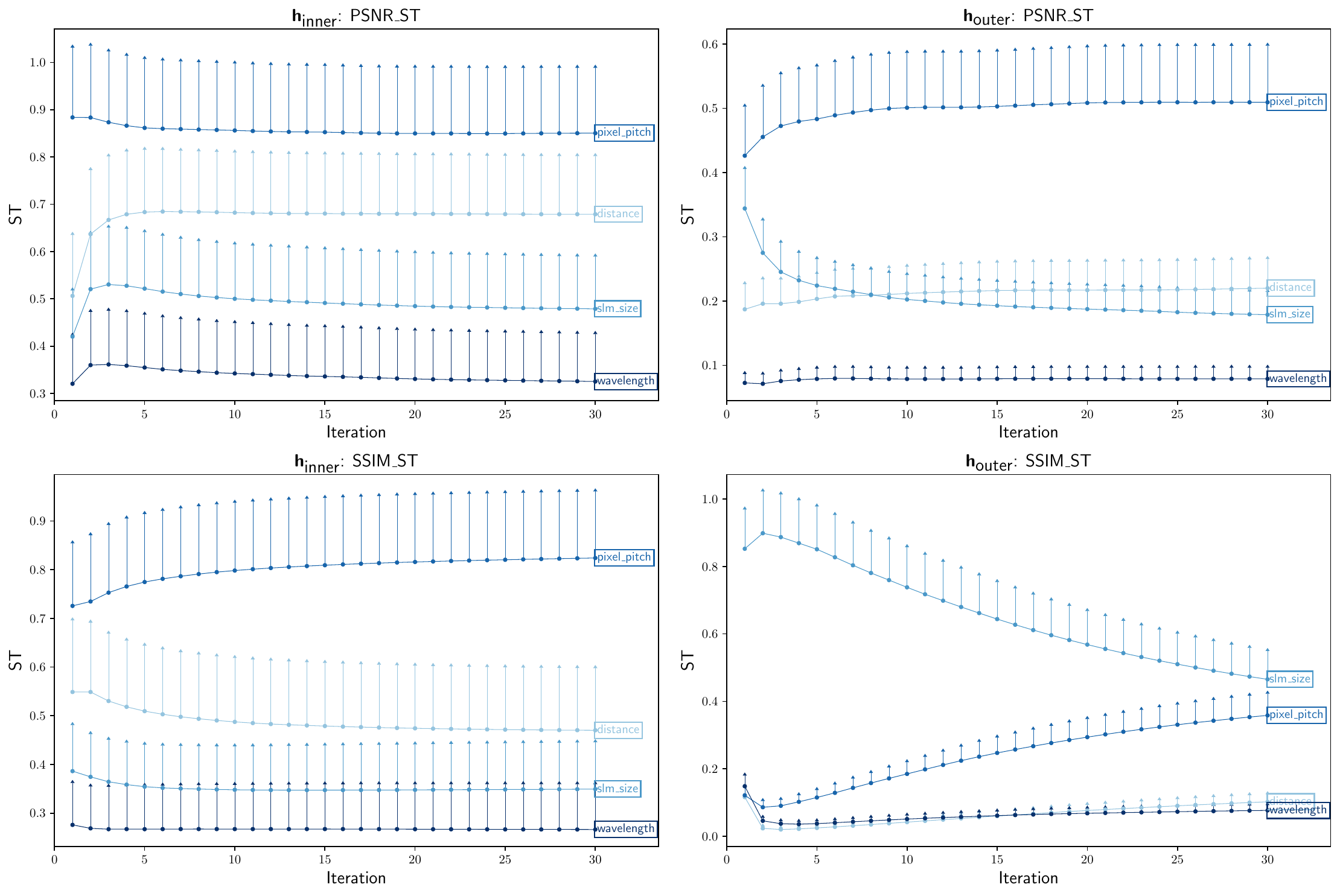}
	\caption{
		Sensitivity Analysis (SA) for $\textbf{\textup{h}}_{\textup{inner}}$ and $\textbf{\textup{h}}_{\textup{outer}}$ (2560 FMH configurations each.) over all iterations for PSNR and SSIM accuracy functions. For SA the accuracy functions were evaluated and averaged over 100 test images for each FMH configuration at each iteration. All panels show the total-order (ST) indices reflecting the overall parameter contributions. The left and right panel correspond to $\textbf{\textup{h}}_{\textup{inner}}$ and $\textbf{\textup{h}}_{\textup{outer}}$ respectively. The top and bottom panels correspond to PSNR and SSIM accuracy functions respectively, as defined in \eqrefp{eq:acuracy_functions}, \eqrefp{eq:acuracy_functions_explain}. Error bars represent the upper limit of the 95\% confidence intervals for clarity.
	}
	\label{fig:0000_25_75_gs_SA_ST_iterations_all}
\end{figure*}

Over multiple iterations, the contributions of certain parameters, such as SLM pixel-resolution and propagation distance, demonstrated perturbations, reflecting potential interaction effects or nonlinear influences. The S1 indices, however, displayed substantial variability within their confidence intervals across parameters, limiting their utility for conclusive analysis.
\paragraph{$\textbf{\textup{h}}_{\textup{inner|outer}}$}
The total-order (ST) sensitivity indices were calculated to assess the influence of input parameters on the GS algorithm performance for both $\textbf{\textup{h}}_{\textup{inner}}$ and $\textbf{\textup{h}}_{\textup{outer}}$, as the first-order (S1) and second-order (S2) indices were found to be unstable due to limited number of experiments.
For $\textbf{\textup{h}}_{\textup{inner}}$ the relative sensitivity ranking for the parameters remain consistent for both the accuracy functions \eqrefp{eq:acuracy_functions} - \eqrefp{eq:acuracy_functions_explain}. The ranking of total-order contributions to the variance in GS algorithm performance was led by pixel-pitch, followed by propagation distance, SLM pixel-resolution, and wavelength. This ranking aligns closely with the results observed for $\textbf{\textup{h}}_{\textup{mid}}$.
For $\textbf{\textup{h}}_{\textup{outer}}$ the relative ST rankings for PSNR accuracy function mirrored those of $\textbf{\textup{h}}_{\textup{inner}}$ and $\textbf{\textup{h}}_{\textup{mid}}$. However, for the SSIM accuracy function, the ranking differed, with SLM pixel-resolution contributing the most, followed by pixel pitch, propagation distance, and wavelength. It is noted that the ST contributions of SLM pixel-resolution decreases and the pixel-pitch increases with the increase in the iterations (\figref{fig:0000_25_75_gs_SA_ST_iterations_all}, \tableref{tab:GS_sa_25_75_PSNR}, \tableref{tab:GS_sa_25_75_SSIM}).
\paragraph{Interaction effects}
Second-order interactions between the parameters were indicated by the sum of the ST indices exceeding 1. However, due to the limited number of experiments, it was not possible to determine which specific parameter interactions contributed most significantly to the variance in the GS algorithm's performance.








\clearpage
\subsection{Forward Model Sensitivity:}

\label{sec:results:sa_fm}
\subsubsection{GS-PINN}

\begin{figure}[H]
	\centering
	\includegraphics[width=3.5in]{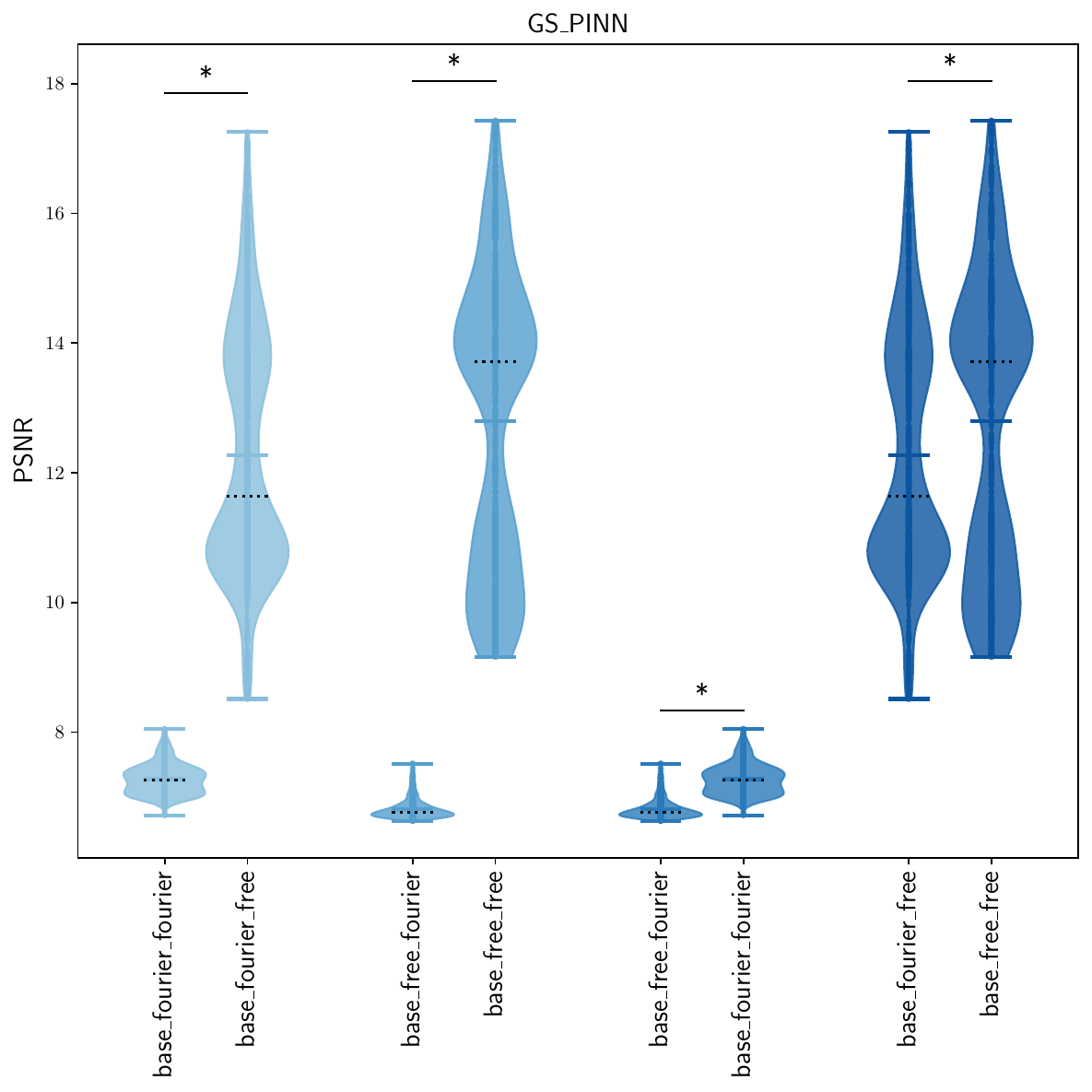}
	\caption{GS-PINN: Forward model comparison with respect to PSNR accuracy. Base models were trained on Fourier holography (base\_fourier) \eqrefp{eq:Fourier_forward_model} or free-space propagation (base\_free) \eqrefp{eq:ASM_forward_model}. Finetuned models (1024 FMH configurations \figref{fig_4}) are labeled as base\_X\_Y, where X indicates the base model and Y the forward model (FM) used for finetuning. Violin plots (medians in dotted black lines, with extremes and mean values) show that free-space propagation consistently outperforms Fourier holography (first two plots), and base models perform better when finetuned with the same FM (last two plots). Wilcoxon signed-rank test (one-sided, alternative: ``less") confirmed significant differences ($p < 0.025$, marked *) include: (i) base\_fourier\_fourier vs. base\_fourier\_free : $W$=$0$, $p$=$2.03\textup{e}^{-169}$, $n$=$1024$, (ii) base\_free\_fourier vs. base\_free\_free : $W$=$0$, $p$=$2.03\textup{e}^{-169}$, $n$=$1024$, (iii) base\_free\_fourier vs base\_fourier\_fourier : $W$=$0$, $p$=$2.03\textup{e}^{-169}$, $n$=$1024$, and (iv) base\_fourier\_free vs. base\_free\_free : $W$=$171,356$, $p$=$3.36\textup{e}^{-22}$, $n$=$1024$.}
	\label{fig:gspinn_free_better_four_violin_psnr}
\end{figure}

\begin{figure*}
	\centering
	\includegraphics[width=6.5in]{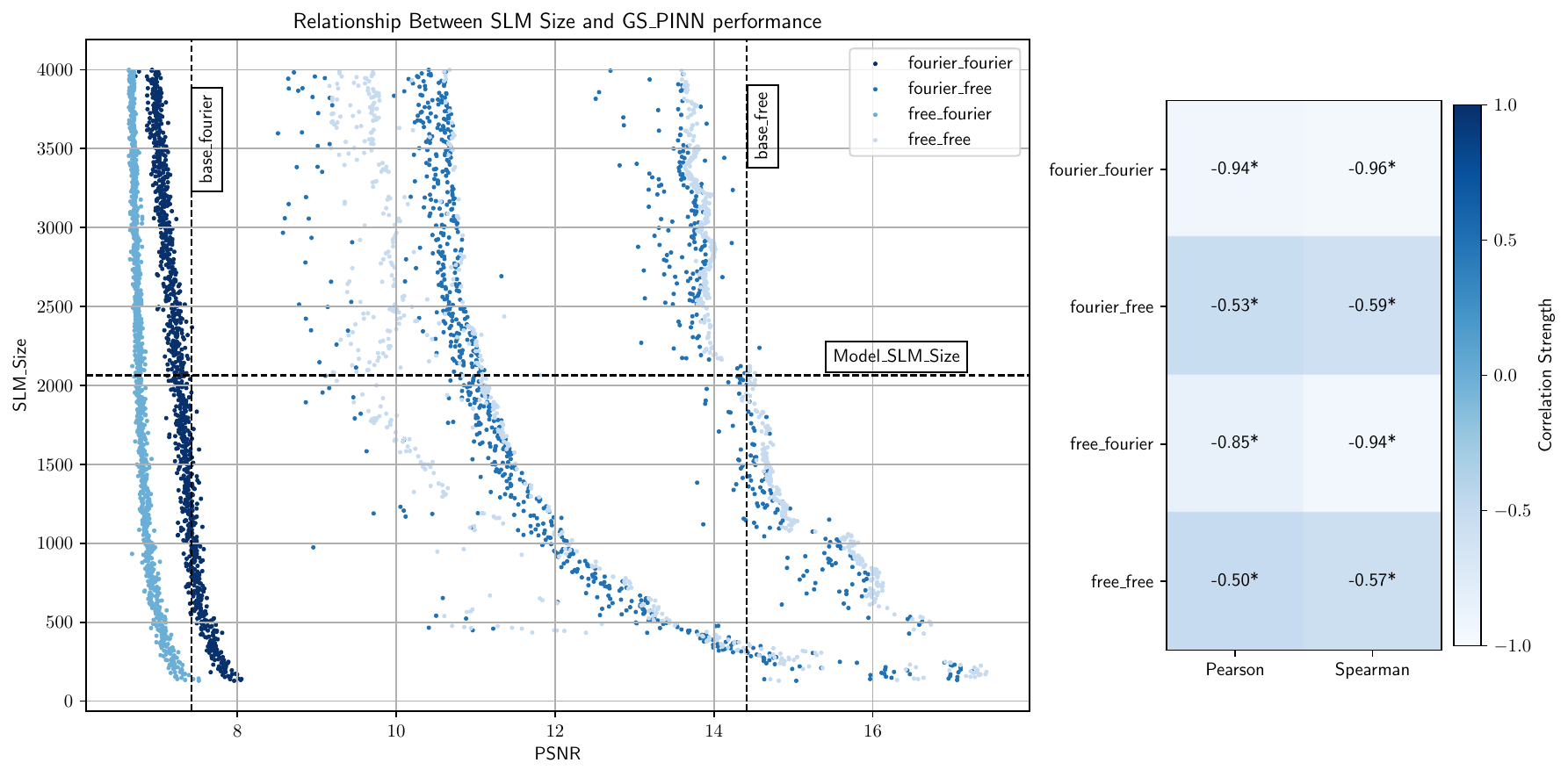}
	\caption{Relationship between SLM pixel-resolution and GS-PINN performance, measured in terms of PSNR accuracy function \eqrefp{eq:acuracy_functions}. Base models were trained on Fourier holography (base\_fourier) \eqrefp{eq:Fourier_forward_model} and free space propagation (base\_free) \eqrefp{eq:ASM_forward_model}. The dotted lines in the left panel indicate the SLM parameters and corresponding PSNR scores for the base models. Finetuned models are labeled as base\_X\_Y, where X denotes the base model and Y specifies the forward model (FM) used for finetuning (1024 FMH configurations \figref{fig_4}). The right panel presents Pearson and Spearman correlation coefficients, with statistically significant correlations ($p < 0.05$) marked by an asterisk (*). Models trained on Fourier holography demonstrate a strong negative correlation between SLM pixel-resolution and PSNR, whereas models trained on free space propagation show weaker correlations.}
	\label{fig:gspinn_corelation_free_four_psnr}
\end{figure*}

\begin{figure}[H]
	\centering
	\includegraphics[width=3.5in]{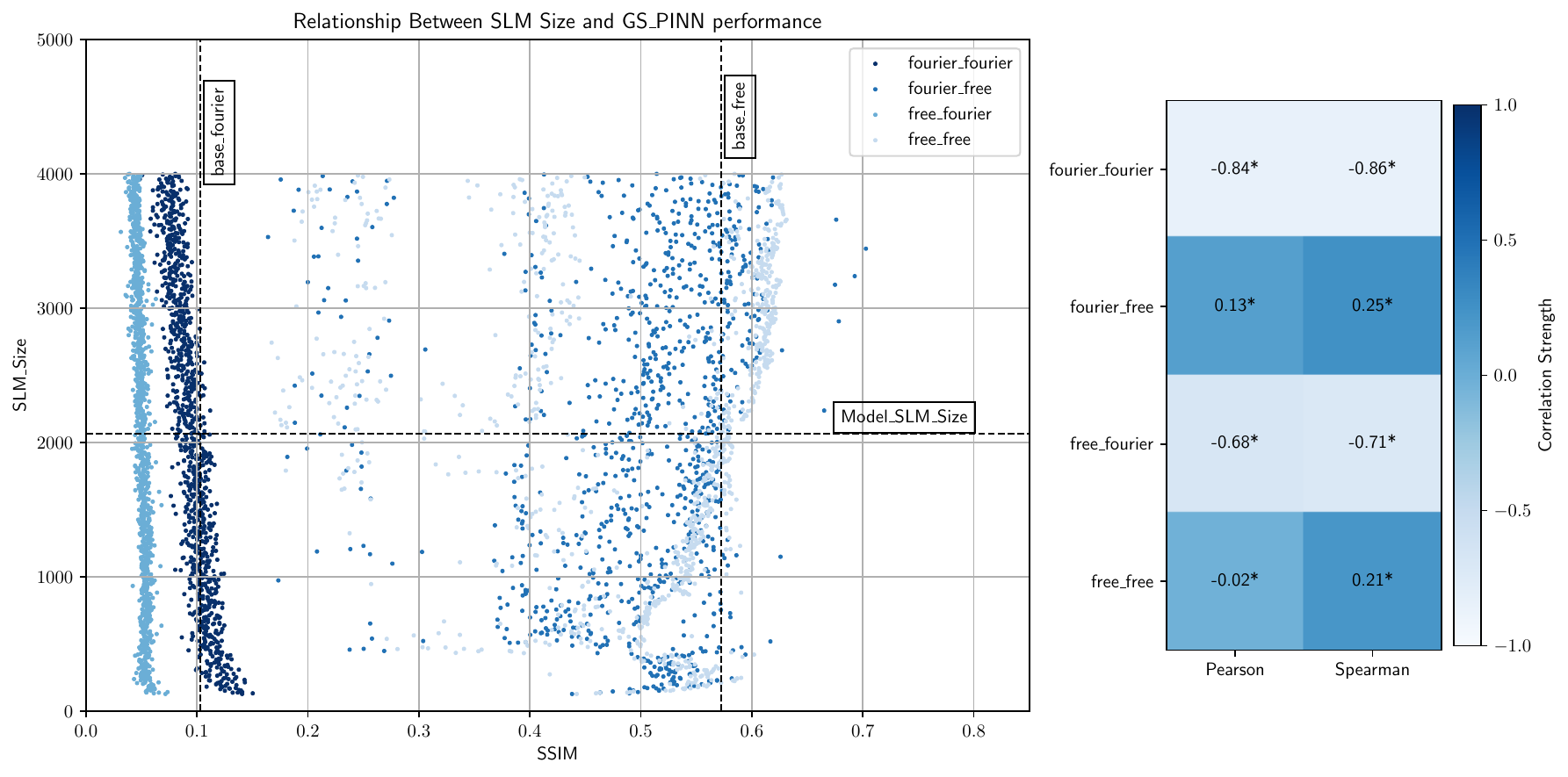}
	\caption{Relationship between Spatial light modulator (SLM) size and GS-PINN performance, measured in terms of SSIM accuracy function \eqrefp{eq:acuracy_functions}. Base models were trained on Fourier holography (base\_fourier) \eqrefp{eq:Fourier_forward_model} and free space propagation (base\_free) \eqrefp{eq:ASM_forward_model}. The dotted lines in the left panel indicate the SLM parameters and corresponding SSIM scores for the base models. Finetuned models are labeled as base\_X\_Y, where X denotes the base model and Y specifies the forward model (FM) (1024 FMH configurations \figref{fig_4}) used for finetuning. The right panel presents Pearson and Spearman correlation coefficients, with statistically significant correlations ($p < 0.05$) marked by an asterisk (*). Models trained on Fourier holography demonstrate a strong negative correlation between SLM pixel-resolution and SSIM, whereas models trained on free space propagation show weaker or negligible correlations.}
	\label{fig:gspinn_corelation_free_four_ssim}
\end{figure}

We applied \algoref[Algorithm 4]{algorithm:finetuning_GS_PINN_FM} to the models $\texttt{base\_fourier}$ and $\texttt{base\_free}$. \figref{fig:gspinn_free_better_four_violin_psnr} and \figref{fig:gspinn_free_better_four_violin_ssim} demonstrates that any base model fine-tuned on free-space propagation consistently outperform those based on Fourier holography. The variance for finetuned base models on free space propagation is larger as compared to Fourier holography. The result remains consistent for both the accuracy functions PSNR and SSIM \eqrefp{eq:acuracy_functions} - \eqrefp{eq:acuracy_functions_explain}. \figref{fig:gspinn_corelation_free_four_psnr} and \figref{fig:gspinn_corelation_free_four_ssim} analyze the relationship between SLM pixel-resolution and GS-PINN performance. For $\texttt{base\_fourier\_fourier}$ finetuned model, a high correlation exists between SLM pixel-resolution and the GS-PINN performance. Finetuning the $\texttt{base\_fourier}$ model for 5 epochs with the free-space propagation model reduces this correlation, resulting in a scatter pattern similar to that of the $\texttt{base\_free\_free}$ finetuned model. Conversely, fine-tuning the $\texttt{base\_free}$ model for 5 epochs on the Fourier holography FM reintroduces high correlation between the SLM pixel-resolution and GS-PINN performance. High correlation relationships, characteristic of Fourier holography, yield poorer GS-PINN performance compared to the low correlation and high interaction effects characteristic in free space propagation.





\subsubsection{GS Algorithm}
\begin{figure}[H]
	\centering
	\includegraphics[width=3.5in]{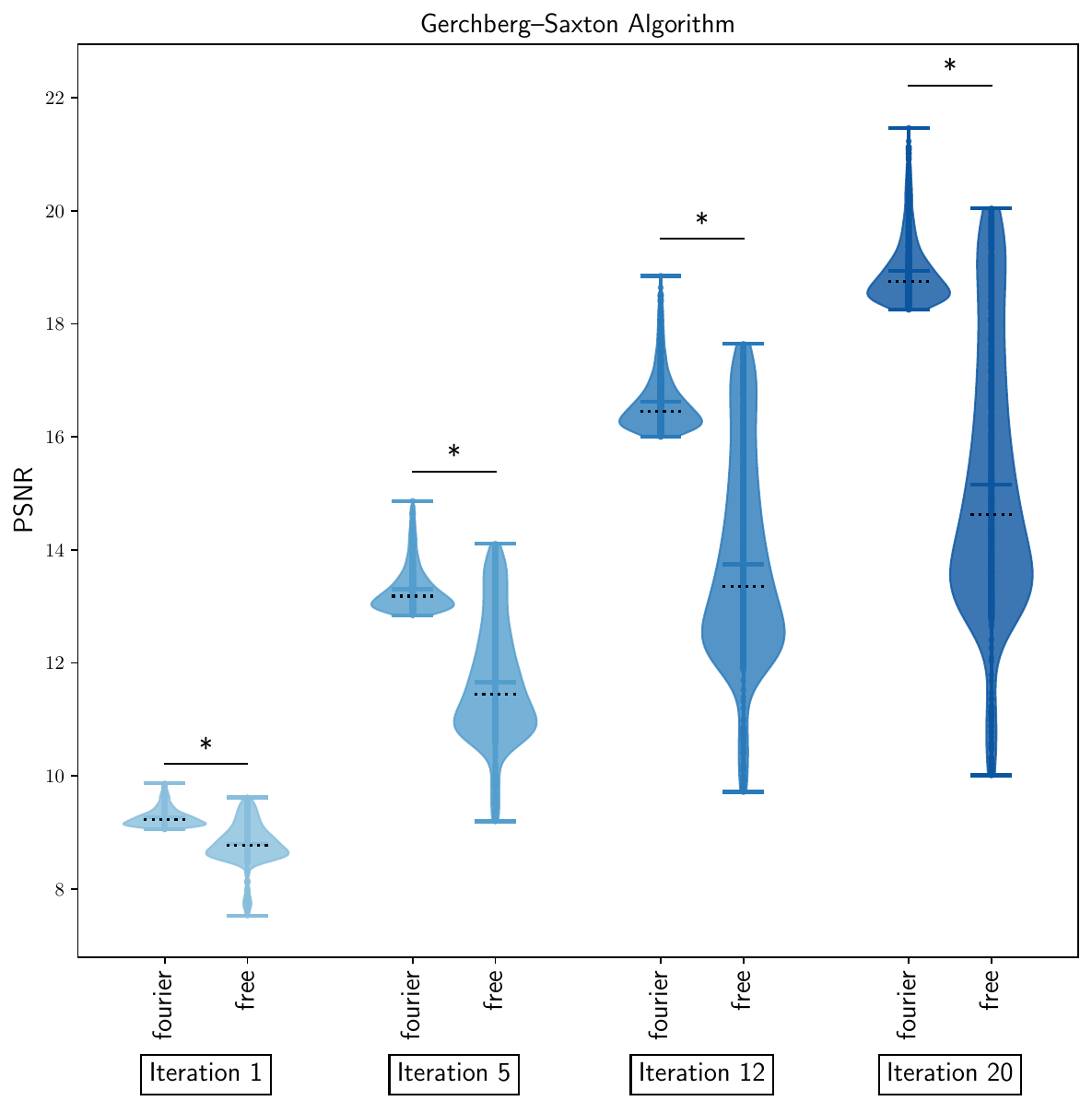}
	\caption{GS algorithm: Forward model comparison (1024 FMH configurations \figref{fig_4}) with respect to PSNR accuracy. Violin plots (medians in dotted black lines, with extremes and mean values) show that Fourier holography \eqrefp{eq:Fourier_forward_model} consistently outperforms free space propagation \eqrefp{eq:ASM_forward_model} for all iterations. Wilcoxon signed-rank test (one-sided, alternative: ``greater") confirmed significant differences ($p < 0.025$, marked *) include: (i) iteration 1 : $W$=$523961$, $p$=$2.36\textup{e}^{-168}$, $n$=$1024$, (ii) iteration 5 : $W$=$524385$, $p$=$6.84\textup{e}^{-169}$, $n$=$1024$, (iii) iteration 12: $W$=$524255$, $p$=$1.0\textup{e}^{-168}$, $n$=$1024$, and (iv) iteration 20 : $W$=$524329$, $p$=$3.36\textup{e}^{-169}$, $n$=$1024$.}
	\label{fig:gs_free_worse_four_violin_psnr}
\end{figure}

\begin{figure*}
	\centering
	\includegraphics[width=6.5in]{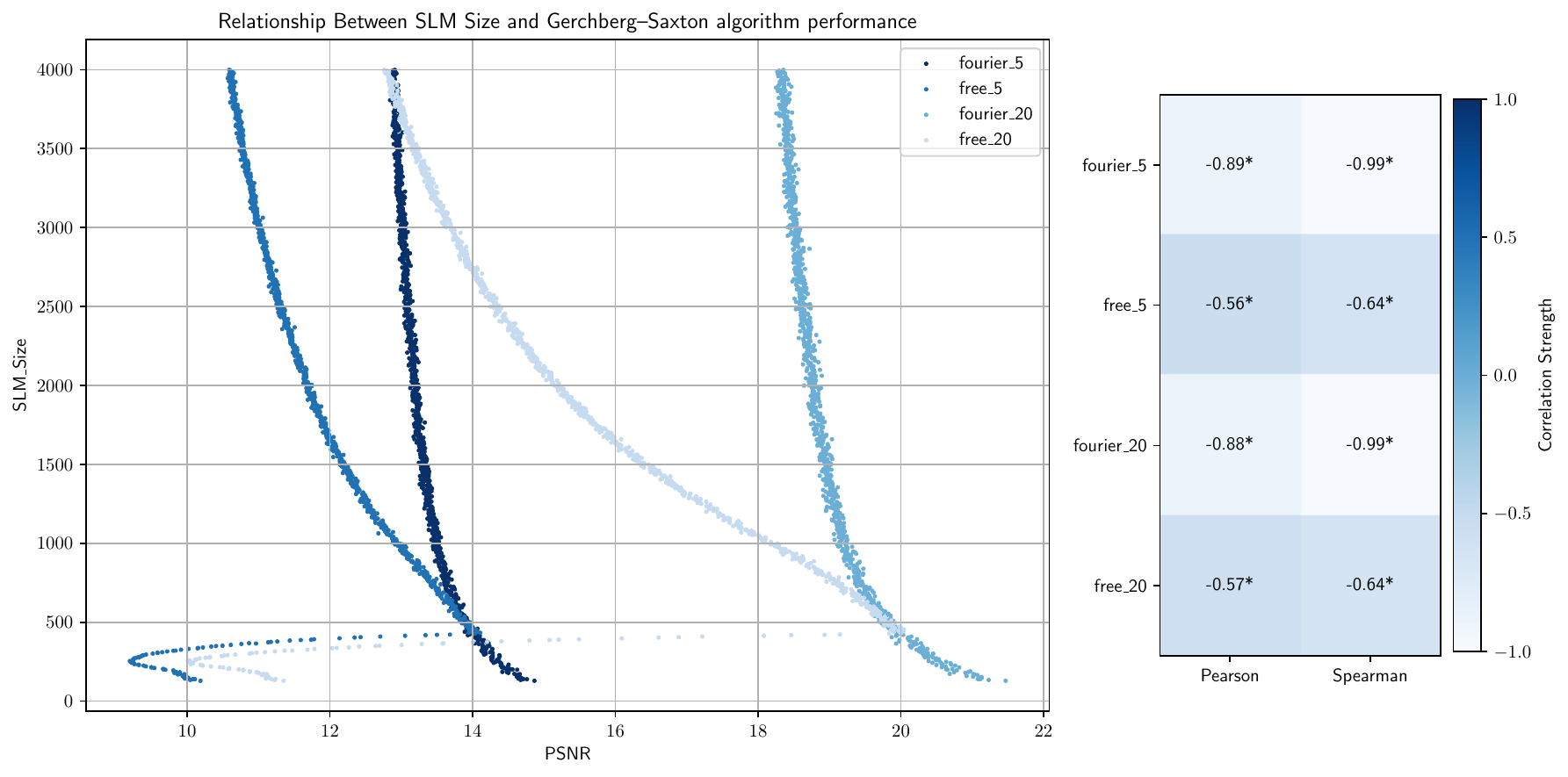}
	\caption{Relationship between SLM pixel-resolution and GS algorithm performance, measured in terms of PSNR accuracy function \eqrefp{eq:acuracy_functions}. Models are labeled as X\_Y, where X denotes the the forward model (FM) (1024 FMH configurations \figref{fig_4}) and Y specifies the iteration number for the GS algorithm. The right panel presents Pearson and Spearman correlation coefficients, with statistically significant correlations ($p < 0.05$) marked by an asterisk (*). Models trained on Fourier holography (fourier\_5|20) \eqrefp{eq:Fourier_forward_model} exhibit a strong negative correlation between SLM pixel-resolution and PSNR, whereas models trained on free space propagation (free\_5|20) \eqrefp{eq:ASM_forward_model} display weaker correlations.}
	\label{fig:gs_corelation_free_four_psnr}
\end{figure*}

\begin{figure}[H]
	\centering
	\includegraphics[width=3.5in]{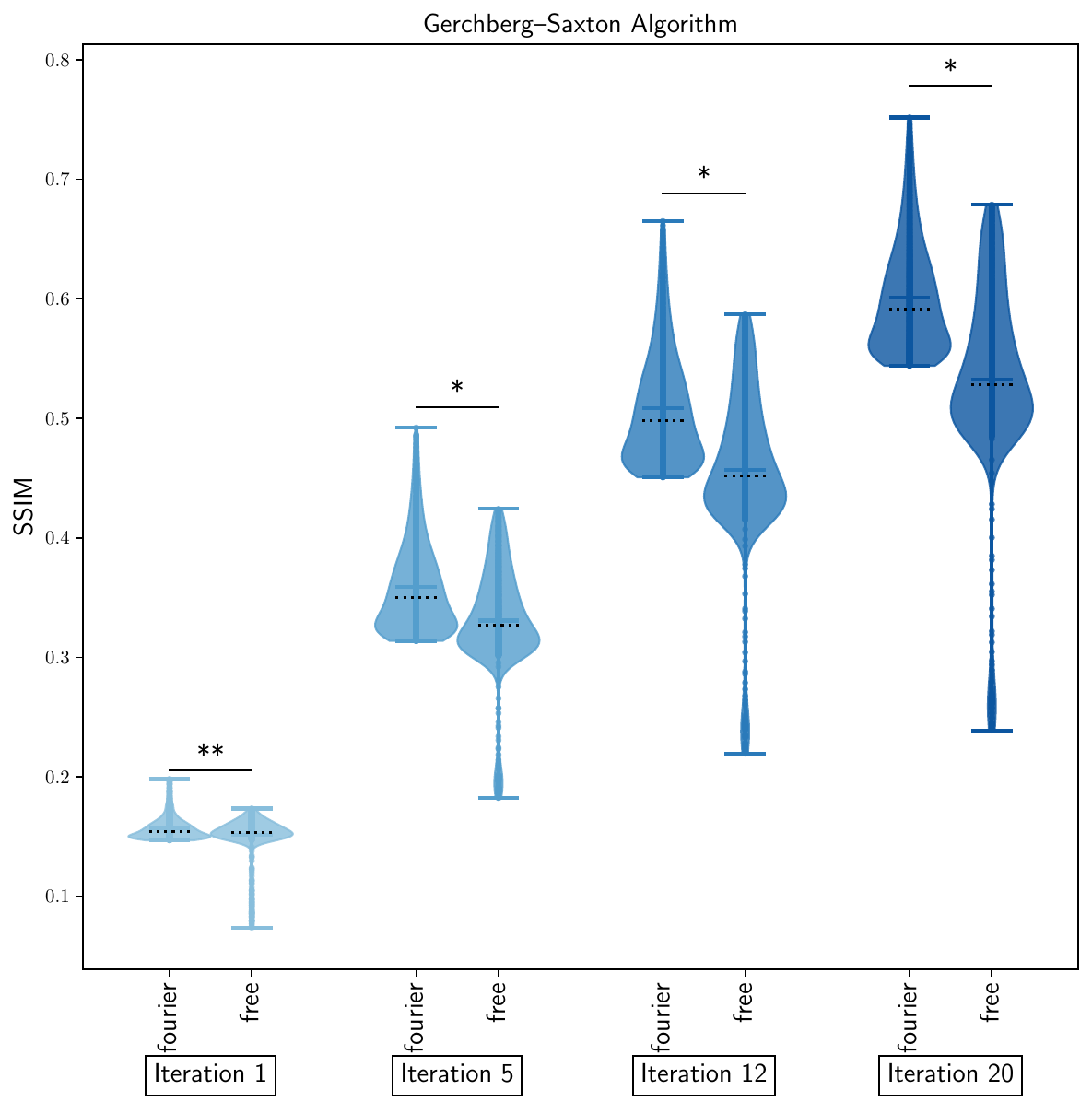}
	\caption{GS algorithm: Forward model comparison (1024 FMH configurations \figref{fig_4}) with respect to SSIM accuracy. Violin plots (medians in dotted black lines, with extremes and mean values) show that Fourier holography \eqrefp{eq:Fourier_forward_model} consistently outperforms free space propagation \eqrefp{eq:ASM_forward_model} for iterations greater than 1. Wilcoxon signed-rank test (one-sided, alternative: ``greater") confirmed significant differences ($p < 0.025$, marked *) include: (i) iteration 5 : $W$=$524539$, $p$=$4.36\textup{e}^{-169}$, $n$=$1024$, (ii) iteration 12: $W$=$524655$, $p$=$3.10\textup{e}^{-169}$, $n$=$1024$, and (iii) iteration 20 : $W$=$524665$, $p$=$3.01\textup{e}^{-169}$, $n$=$1024$. For the first iteration Fourier holography was not significantly ($p > 0.025$, marked **) better than free space propagation (iteration 1 : $W$=$159466$, $p$=$1.0$, $n$=$1024$)}
	\label{fig:gs_free_worse_four_violin_ssim}
\end{figure}

\begin{figure}[H]
	\centering
	\includegraphics[width=3.5in]{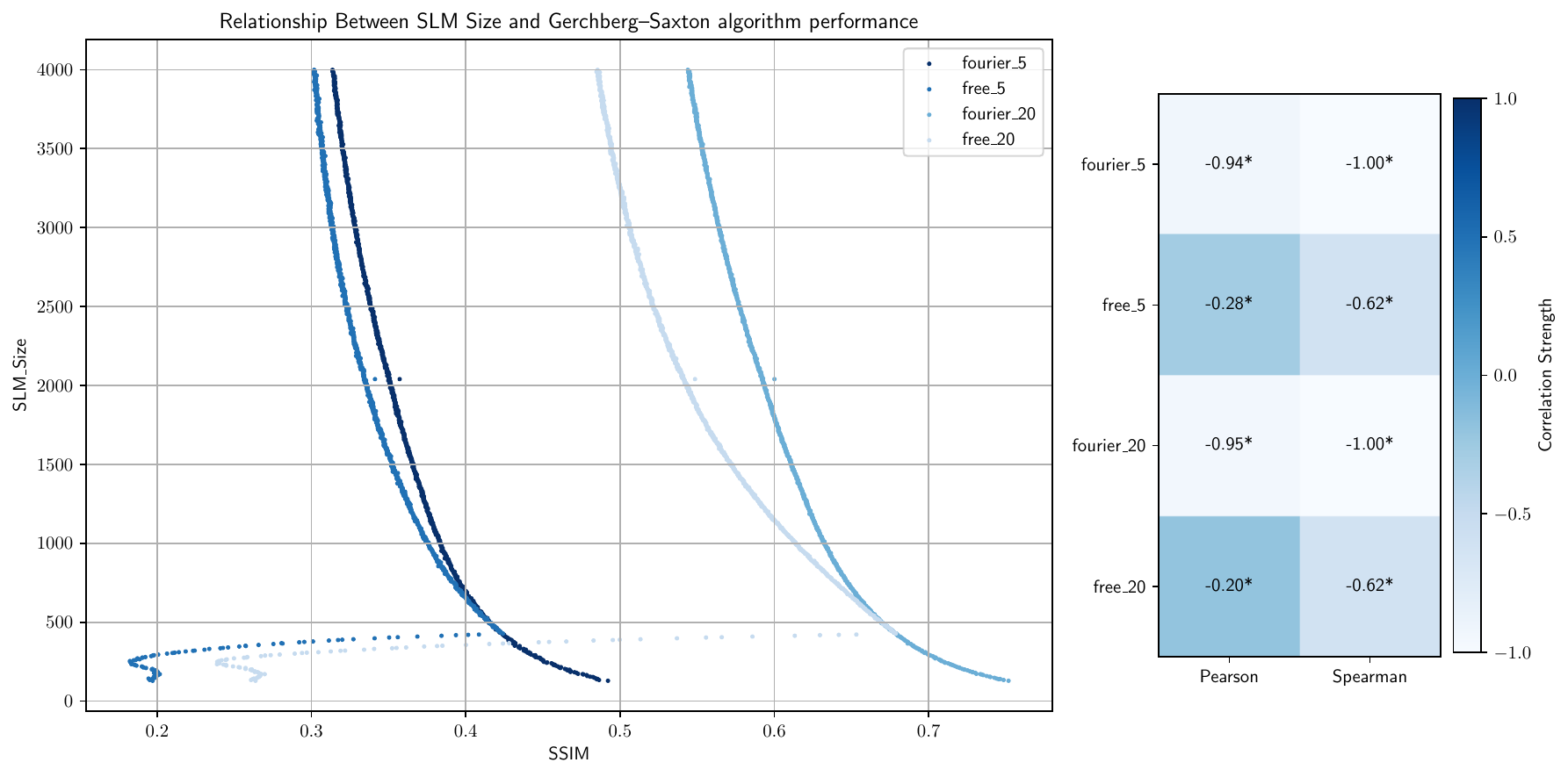}
	\caption{Relationship between Spatial light modulator (SLM) size and GS algorithm performance, measured in terms of SSIM accuracy function \eqrefp{eq:acuracy_functions}. Models are labeled as X\_Y, where X denotes the the forward model (FM) (1024 FMH configurations \figref{fig_4}) and Y specifies the iteration number for the GS algorithm. The right panel presents Pearson and Spearman correlation coefficients, with statistically significant correlations ($p < 0.05$) marked by an asterisk (*). Models trained on Fourier holography (fourier\_5|20) \eqrefp{eq:Fourier_forward_model} exhibit a strong negative correlation between SLM pixel-resolution and PSNR, whereas models trained on free space propagation (free\_5|20) \eqrefp{eq:ASM_forward_model} display weaker correlations.}
	\label{fig:gs_corelation_free_four_ssim}
\end{figure}

We conducted a similar analysis for the GS algorithm as performed for GS-PINN by using \algoref[Algorithm 6]{algorithm:finetuning_GS_FM}. \figref{fig:gs_free_worse_four_violin_psnr} and \figref{fig:gs_free_worse_four_violin_ssim} show that the GS algorithm evaluated on Fourier holography outperforms free-space propagation when the same FMH configurations are used. This result holds across increasing iterations and different accuracy functions \eqrefp{eq:acuracy_functions} - \eqrefp{eq:acuracy_functions_explain}. The variance in performance is notably higher for GS models based on free-space propagation compared to Fourier holography. \figref{fig:gs_corelation_free_four_psnr} and \figref{fig:gs_corelation_free_four_ssim} examine the relationship between SLM pixel-resolution and the GS algorithm's performance. For Fourier holography, there is a significant correlation between SLM pixel-resolution and performance, persisting through the 5th and 20th iterations. In contrast, the correlation diminishes for GS models evaluated on free-space propagation. These results are the inverse of those observed for GS-PINN, highlighting distinct performance dynamics between the two approaches.

Refer to \secref{supplemetary:sec:D}  for further results and caveats.

\clearpage

\clearpage

\subsection{Composite metric:}
\label{sec:results:composite_metric}
\begin{figure}[H]
	\centering
	\includegraphics[width=3.5in]{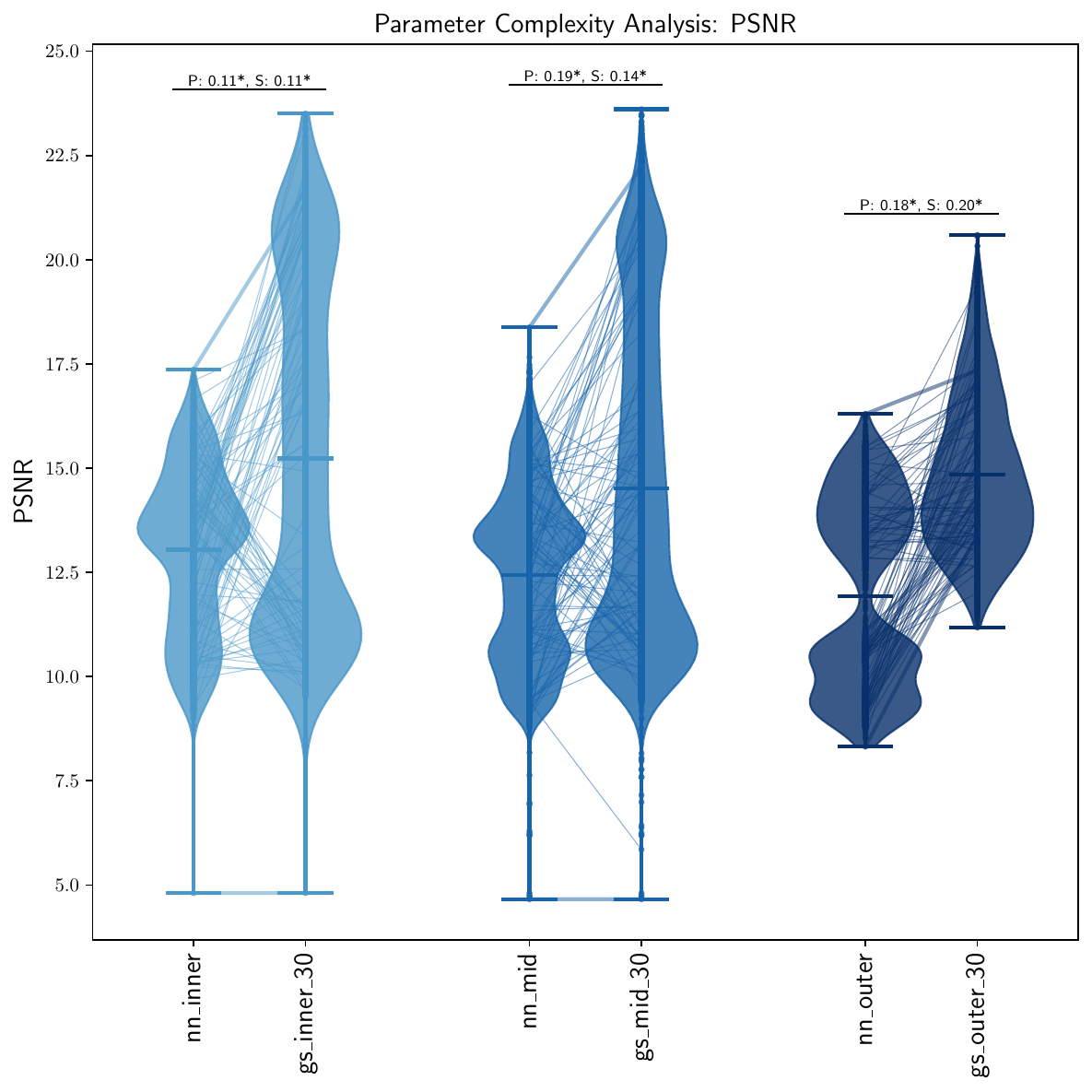}
	\caption{Parameter complexity analysis for GS-PINN and the GS algorithm, evaluated using the PSNR accuracy function \eqrefp{eq:acuracy_functions}. Models are labeled as X\_Y, where X represents either the GS-PINN (nn) or GS algorithm (gs), and Y corresponds to configurations of $\textbf{\textup{h}}_{\textup{inner}}$ (inner), $\textbf{\textup{h}}_{\textup{mid}}$ (mid), and $\textbf{\textup{h}}_{\textup{outer}}$ (outer). For the GS algorithm, the $30^{\textup{th}}$  iteration is shown. Both GS-PINN and GS algorithm models were trained on similar FMH configurations: 10,240 FMH configurations for $\textbf{\textup{h}}_{\textup{mid}}$ and 2,560 FMH configurations each for $\textbf{\textup{h}}_{\textup{outer}}$ and $\textbf{\textup{h}}_{\textup{inner}}$. Violin plots depict the mean and extremes of the distributions. Pearson (P) and Spearman (S) correlation coefficients are displayed at the top of each pair of violin plots, with statistically significant correlations ($p < 0.025$) marked by an asterisk (*) and non-significant correlations ($p > 0.025$) denoted by a double asterisk (**). Results indicate that both GS-PINN and GS algorithm models exhibit weak to negligible statistically significant correlations in performance when trained on similar FMH configurations, as measured by the PSNR accuracy function.}
	\label{parameter_complexity_analysis_psnr}
\end{figure}

\begin{figure}
	\centering
	\includegraphics[width=3.5in]{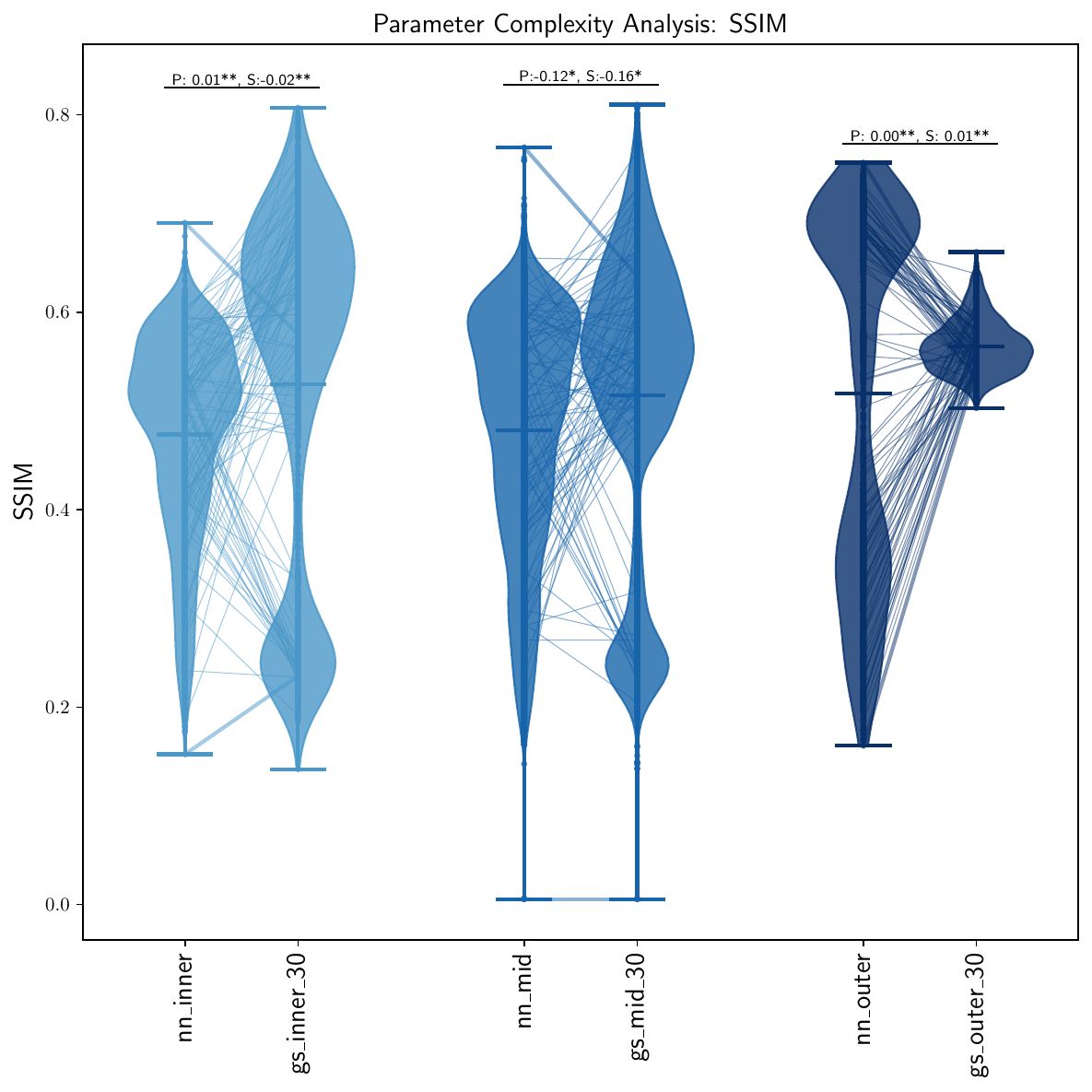}
	\caption{Parameter complexity analysis for GS-PINN and the GS algorithm, evaluated using the SSIM accuracy function \eqrefp{eq:acuracy_functions}. Models are labeled as X\_Y, where X represents either the GS-PINN (nn) or GS algorithm (gs), and Y corresponds to configurations of $\textbf{\textup{h}}_{\textup{inner}}$ (inner), $\textbf{\textup{h}}_{\textup{mid}}$ (mid), and $\textbf{\textup{h}}_{\textup{outer}}$ (outer). For the GS algorithm, the $30^{\textup{th}}$ iteration is shown. Both GS-PINN and GS algorithm models were trained on similar FMH configurations: 10,240 FMH configurations for $\textbf{\textup{h}}_{\textup{mid}}$ and 2,560 FMH configurations each for $\textbf{\textup{h}}_{\textup{outer}}$ and $\textbf{\textup{h}}_{\textup{inner}}$. Violin plots depict the mean and extremes of the distributions. Pearson (P) and Spearman (S) correlation coefficients are displayed at the top of each pair of violin plots, with statistically significant correlations ($p < 0.025$) marked by an asterisk (*) and non-significant correlations ($p > 0.025$) denoted by a double asterisk (**). For $\textbf{\textup{h}}_{\textup{outer}}$ and $\textbf{\textup{h}}_{\textup{inner}}$ configurations, weak and non-significant correlations suggest no meaningful linear or monotonic relationship between the SSIM values of GS-PINN and GS algorithm models. The performances of the models in these configurations are largely independent. However, for the $\textbf{\textup{h}}_{\textup{mid}}$ configuration, weak but statistically significant negative correlations are observed, indicating a slight inverse relationship between the performances of the two models.}
	\label{parameter_complexity_analysis_ssim}
\end{figure}

Here we analyze if the complexity experienced by the GS-PINN correlates to that of GS algorithm for similar FMH's (\figref{parameter_complexity_analysis_psnr}, \figref{parameter_complexity_analysis_ssim}). Here ``nn" and ``gs'' correspond to GS-PINN and GS algorithm respectively. ``inner", ``mid", ``outer" correspond to $\textbf{\textup{h}}_{\textup{inner}}$, $\textbf{\textup{h}}_{\textup{mid}}$,  $\textbf{\textup{h}}_{\textup{outer}}$ respectively. $\textbf{\textup{h}}_{\textup{inner}}$, $\textbf{\textup{h}}_{\textup{mid}}$,  $\textbf{\textup{h}}_{\textup{outer}}$ correspond to 2560, 10240, 2560 different configurations of FMH around the points according to the bounds \tableref{table:inner_mid_outer_points} and \eqrefp{eq:FMH_bounds}. Each point in the violin plot corresponds to a single FMH model. The corresponding PSNR and SSIM scores are the average $\overline{\textup{PSNR}}$ and $\overline{\textup{SSIM}}$ calculated after evaluating the models on 100 test images from the test data. Pearson and Spearman correlation coefficients were calculated between the relevant pairs ($\textbf{\textup{h}}_{\textup{inner}}$, $\textbf{\textup{h}}_{\textup{mid}}$,  $\textbf{\textup{h}}_{\textup{outer}}$) of GS algorithm and GS-PINN performance. Both coefficients were found to be near zero, indicating a very weak or negligible correlation between the complexity of FMH associated with the GS algorithm and that of the GS-PINN. Moreover, the parameter complexity analysis remained consistent across the accuracy functions defined in \eqrefp{eq:acuracy_functions} - \eqrefp{eq:acuracy_functions_explain}, reinforcing the observation of minimal correlation between the two methods.

\section{Discussion}
\label{sec:discussion}
This study aimed to explore the sensitivity of forward model hyperparameters (FMHs) and forward models (FMs) on the performance of both the GS algorithm and GS-PINN. By quantifying FMH sensitivity, evaluating FM performance, and benchmarking metrics, the work offers valuable insights for optimizing holographic systems and fostering experimental and theoretical CGH research.
\subsection{Influence of FMHs on System Performance:}
(\secref{sec:results:sa_fmh}) Our results demonstrate that SLM pixel-resolution is the most influential parameter for GS-PINN across various experiments. Pixel-pitch emerges as the second most significant contributor, with smaller effects observed for propagation distance and wavelength.  Interestingly, the relative sensitivity rankings for SLM pixel-resolution and pixel-pitch remain consistent across different accuracy functions, such as PSNR and SSIM, for $\textbf{\textup{h}}_{\textup{inner}}$, $\textbf{\textup{h}}_{\textup{mid}}$ and $\textbf{\textup{h}}_{\textup{outer}}$. This suggests that, regardless of the accuracy metric used, SLM pixel-resolution and pixel-pitch have a pronounced impact on the overall system performance.
In contrast, for the GS algorithm, the sensitivity analysis reveals that pixel-pitch dominates the variance in $\textbf{\textup{h}}_{\textup{inner}}$, $\textbf{\textup{h}}_{\textup{mid}}$ and $\textbf{\textup{h}}_{\textup{outer}}$ for both accuracy metrics. Eventhough in  $\textbf{\textup{h}}_{\textup{outer}}$ the sensitivity between SLM pixel-resolution and pixel-pitch is switched, it was evident that the sensitivity of pixel-pitch was continuously increasing and SLM-size decresing with the increase in iterations. Perturbations in the evolution of sensitivity across multiple iterations in the GS algorithm suggests that the sensitivity of these parameters evolves non-linearly over iterations, potentially due to interaction effects or changing parameter dependencies.
Notably, in both cases, GS-agorithm and GS-PINN, interactions between SLM-related parameters (SLM pixel-resolution and pixel-pitch) were stronger than those involving optical parameters (wavelength and propagation distance). This observation emphasizes the crucial role of hardware-driven FMHs, particularly SLM-related parameters over optical parameters in determining system performance. These findings suggest that in optimizing holographic systems, priority should be given to parameters that are hardware-dependent, especially in systems constrained by hardware limitations.
\subsection{Influence of FMs on System Performance:}
(\secref{sec:results:sa_fm}) For GS-PINN, free space propagation consistently outperformed Fourier holography, demonstrating superior overall performance in both PSNR and SSIM accuracy metrics. While free space propagation exhibited higher variance, it also showed reduced correlation strength between SLM pixel-resolution and performance, indicating greater flexibility and generalization potential. This suggests a tradeoff between performance stability and the ability to generalize, guiding neural network training decisions. In contrast, Fourier holography performed better with the GS algorithm, exhibiting lower variance and stronger correlations between SLM pixel-resolution and performance.
The divergent trends between the two models highlight the importance of choosing the appropriate forward model based on the specific algorithm and performance goals. Overall, our findings demonstrate that free space propagation offers advantages in generalization for GS-PINN, while Fourier holography provides stability for the GS algorithm. This comparison helps experimentalists select the most suitable forward model for neural network-based holographic systems, emphasizing the need for model selection based on the specific goals of the optimization process.
\subsection{Impact of FMH-Dependent Complexity on Algorithm Benchmarking:}
\label{sec:discussion:composite_metric}

Building on insights into FMHs and FMs, we sought to assess the validity of benchmarking new algorithms against the traditional iterative GS algorithm or other neural networks. Our analysis, shown in \figref{parameter_complexity_analysis_psnr}, \figref{parameter_complexity_analysis_ssim}, reveals that the FMH-associated complexity experienced by GS-PINN does not significantly correlate with that of the GS algorithm. Both Pearson and Spearman correlation coefficients were near zero, indicating negligible linear and rank-based relationships between the two methods.
This lack of correlation underscores distinct complexity patterns between GS-PINN and the GS algorithm when handling similar FMH configurations. These findings suggest that comparing neural networks trained on one FMH configuration with those trained on different FMHs (or with the GS algorithm) could lead to unfair evaluations, as the complexity encountered during training varies across configurations. Our observation is based on a single GS-PINN variant using the initialization network. We hypothesize that this behavior will remain consistent across other GS-PINN variants, as well as the GS algorithm. This hypothesis is supported by the fact that CGH is an ill-posed problem, and all unsupervised GS-PINN variants can be considered unrolled versions of the GS algorithm. Thus, we expect similar behavior across other variants. However, providing a robust mathematical proof or conducting additional experiments with other variants to substantiate this hypothesis is beyond the scope of this work.
To address these challenges, we developed a composite metric that combines the GS-weighted metric, generalization metric, and resilience metric. This metric ensures reliable evaluations, avoids speculative conclusions, and provides a standardized framework for comparing performance across diverse algorithms and FMH configurations.

\subsection{Impact of FMH Sensitivity on Model Interpretability:}
The analysis from \secref{sec:results:sa_fmh} reveals that the phase initialization network within GS-PINN (\figref{fig:GS_PINN}) effectively abstracts the influence of pixel-pitch, a FMH that contributes the most to the variance in GS algorithm performance. Instead, the network relies on SLM pixel-resolution, which becomes the dominant contributor to GS-PINN's performance. Furthermore, GS-PINN minimizes the impact of variables associated with the optical parameters of the system, highlighting its ability to focus on hardware-related aspects. This informed analysis not only enhances model interpretability but also paves the way for developing more generalized and explainable networks. Such networks provide clearer insights into how perturbations in specific parameters affect performance, aiding in the creation of targeted networks. In line with our initial objectives, this work enhances model interpretability and generalization by identifying key FMHs that influence network performance. This facilitates more informed interpretations of outputs, supporting the development of AI models that are potentially more explainable and adaptable to various holographic applications.
\subsection{Limitations:}
In this study, we trained the networks on a single FMH configuration corresponding to $\textbf{\textup{h}}_{\textup{inner}}$, $\textbf{\textup{h}}_{\textup{mid}}$ and $\textbf{\textup{h}}_{\textup{outer}}$ and subsequently fine-tuned the models for the remaining FMH configurations as outlined in \tableref{table:inner_mid_outer_points} and \eqrefp{eq:FMH_bounds}, with evaluation of FM sensitivity based on the configurations in \figref{fig_4}. However, due to computational constraints, we did not train separate networks from scratch for each FMH configuration. Additionally, the analysis was conducted using a single variant of the GS-PINN framework. Future work could explore the use of alternative GS-PINN variants with different network architectures, which may provide deeper insights into the impact of network design on performance and sensitivity.
\section{Conclusion}
\label{sec:conclusion}
In this work, we proposed a structured framework for evaluating neural networks in the context of computer-generated holography (CGH), with a particular focus on the influence of forward model hyperparameters (FMHs) and forward models (FMs). We further examined the impact of FMH-dependent complexity on algorithm benchmarking and explored how FMH-dependent sensitivity affects model interpretability and generalization. By providing a comprehensive methodology for selecting appropriate forward models, hyperparameters, and evaluation metrics, this study contributes to informed decision-making in both experimental and theoretical CGH research. Future work may build upon these findings by investigating additional variants of GS-PINN and assessing their performance across a wider range of forward models and configurations, enabling further advancements in CGH-related neural network applications.

\bibliographystyle{IEEEtran}
\bibliography{sa}

\clearpage

\section{Supplementary}
\subsection{Forward model sensitivity of GS-PINN and GS algorithm for SSIM accuracy function.}
\begin{figure}[H]
	\centering
	\includegraphics[width=3.5in]{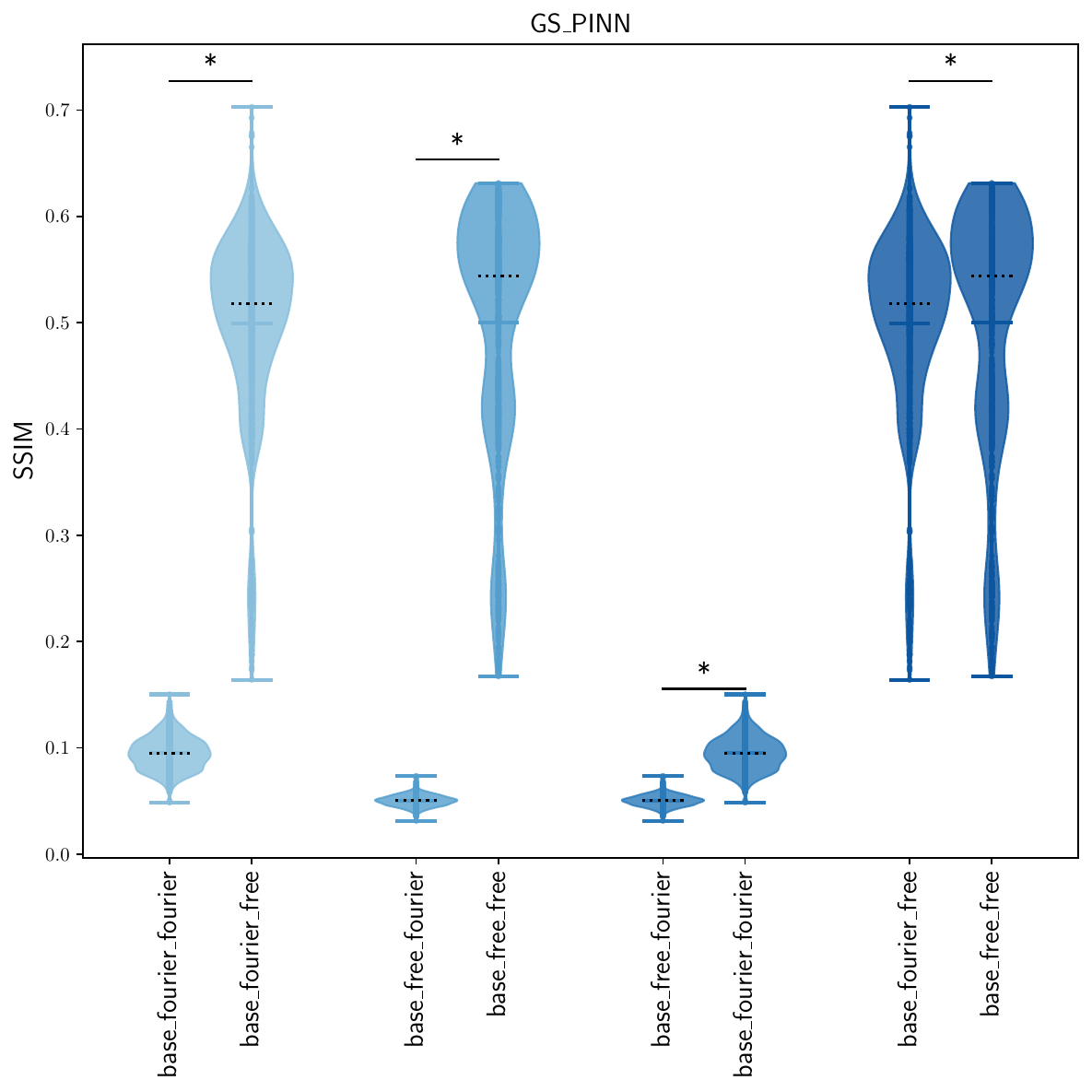}
	\caption{GS-PINN: Forward model comparison with respect to SSIM accuracy. Base models were trained on Fourier holography (base\_fourier) \eqrefp{eq:Fourier_forward_model} or free space propagation (base\_free) \eqrefp{eq:ASM_forward_model}. Finetuned models are labeled as base\_X\_Y, where X indicates the base model and Y the forward model (FM) (1024 FMH configurations \figref{fig_4}) used for finetuning. Violin plots (medians in dotted black lines, with extremes and mean values) show that free space propagation consistently outperforms Fourier holography (first two plots), and base models perform better when finetuned with the same FM (last two plots). Wilcoxon signed-rank test (one-sided, alternative: ``less") confirmed significant differences ($p < 0.025$, marked *) include: (i) base\_fourier\_fourier vs. base\_fourier\_free : $W$=$0$, $p$=$2.03\textup{e}^{-169}$, $n$=$1024$, (ii) base\_free\_fourier vs. base\_free\_free : $W$=$0$, $p$=$2.03\textup{e}^{-169}$, $n$=$1024$, (iii) base\_free\_fourier vs base\_fourier\_fourier\ : $W$=$0$, $p$=$2.03\textup{e}^{-169}$, $n$=$1024$, and (iv) base\_fourier\_free vs. base\_free\_free : $W$=$223023$, $p$=$1.59\textup{e}^{-5}$, $n$=$1024$.}
	\label{fig:gspinn_free_better_four_violin_ssim}
\end{figure}

\clearpage
\subsection{Algorithms for SA of forward model hyperparameters and forward models for GS algorithm.}
\begin{figure}[H]
	\centering
        \includegraphics[height=6in]{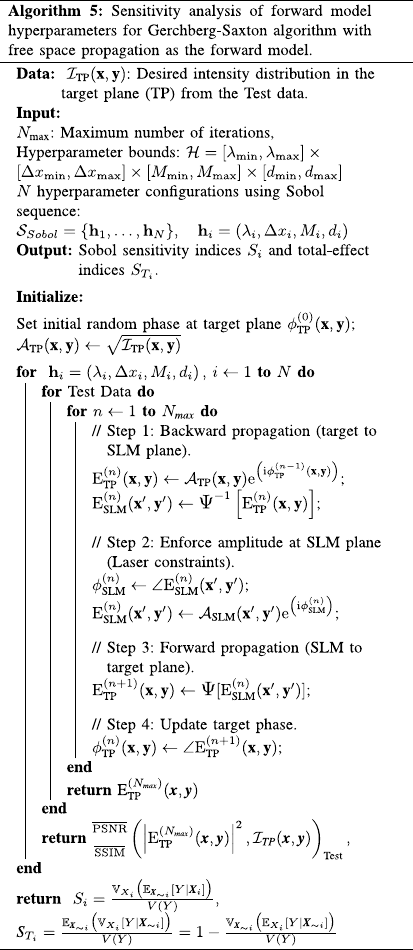}
	\caption*{}
	\label{algorithm:finetuning_GS_FMH}
\end{figure}

\begin{figure}[H]
	\centering
        \includegraphics[height=6in]{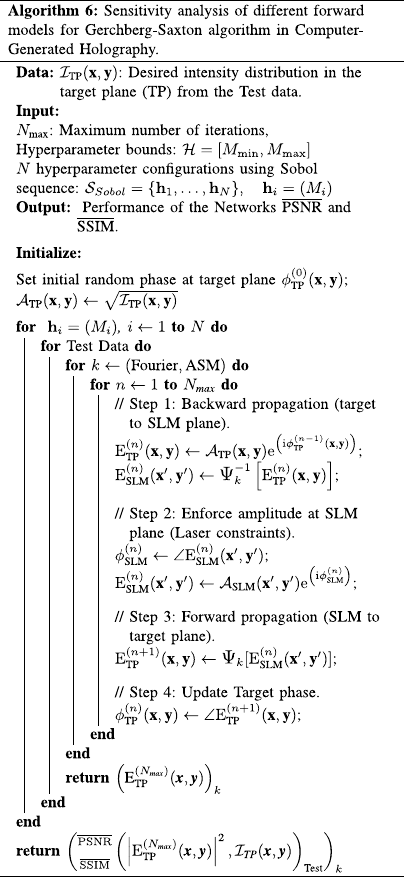}
	\caption*{}
	\label{algorithm:finetuning_GS_FM}
\end{figure}

\clearpage
\subsection{Absolute sensitivity indices for GS-PINN and GS algorithms for different accuracy functions across all the FMH experiments.}
\begin{table*}[b]
	\caption{PSNR: GS-PINN sensitivity indices (ST, S1, S2) for $\textbf{\textup{h}}_{\textup{mid}}.$}
	\label{tab:GSPINN_sa_50_PSNR}
	\begin{center}
		\begin{tabular}{|l|c|c|c|}
			\hline
			$\textbf{\textup{Parameter}}$      & $\textbf{\textup{ST}} \pm \textbf{\textup{ST\_conf}}$ & $\textbf{\textup{S1}} \pm \textbf{\textup{S1\_conf}}$ & $\textbf{\textup{S2}} \pm \textbf{\textup{S2\_conf}}$ \\
			\hline
			$\textup{distance} (d)$            & $0.190517 \pm 0.025090$                               & $0.027013 \pm 0.040998$                               & -                                                     \\
			$\textup{SLM\_size} (M)$           & $0.820431 \pm 0.083134$                               & $0.565332 \pm 0.077603$                               & -                                                     \\
			$\textup{pixel\_pitch} (\Delta x)$ & $0.367135 \pm 0.042765$                               & $0.065698 \pm 0.054043$                               & -                                                     \\
			$\textup{wavelength} (\lambda)$    & $0.197639 \pm 0.029199$                               & $-0.007307 \pm 0.036396$                              & -                                                     \\
			\hline
			$(d, M)$                           & -                                                     & -                                                     & $0.015795 \pm 0.058254$                               \\
			$(d, \Delta x)$                    & -                                                     & -                                                     & $0.014212 \pm 0.048451$                               \\
			$(d, \lambda)$                     & -                                                     & -                                                     & $0.007966 \pm 0.051702$                               \\
			$(M, \Delta x)$                    & -                                                     & -                                                     & $0.066788 \pm 0.081633$                               \\
			$(M, \lambda)$                     & -                                                     & -                                                     & $0.012854 \pm 0.080150$                               \\
			$(\Delta x, \lambda)$              & -                                                     & -                                                     & $0.031972 \pm 0.064039$                               \\
			\hline
		\end{tabular}
	\end{center}
\end{table*}

\begin{table*}[b]
	\caption{SSIM: GS-PINN sensitivity indices (ST, S1, S2) for $\textbf{\textup{h}}_{\textup{mid}}.$}
	\label{tab:GSPINN_sa_50_SSIM}
	\begin{center}
		\begin{tabular}{|l|c|c|c|}
			\hline
			$\textbf{\textup{Parameter}}$      & $\textbf{\textup{ST}} \pm \textbf{\textup{ST\_conf}}$ & $\textbf{\textup{S1}} \pm \textbf{\textup{S1\_conf}}$ & $\textbf{\textup{S2}} \pm \textbf{\textup{S2\_conf}}$ \\
			\hline
			$\textup{distance} (d)$            & $0.398741\pm 0.049578$                                & $0.028386 \pm 0.056150$                               & -                                                     \\
			$\textup{SLM\_size} (M)$           & $0.957969 \pm 0.081744$                               & $0.429896 \pm 0.072036$                               & -                                                     \\
			$\textup{pixel\_pitch} (\Delta x)$ & $0.479556 \pm 0.064865$                               & $0.035397 \pm 0.057683$                               & -                                                     \\
			$\textup{wavelength} (\lambda)$    & $0.377307 \pm 0.048125$                               & $-0.014509 \pm 0.053587$                              & -                                                     \\
			\hline
			$(d, M)$                           & -                                                     & -                                                     & $0.064046 \pm 0.086028$                               \\
			$(d, \Delta x)$                    & -                                                     & -                                                     & $0.026214 \pm 0.078142$                               \\
			$(d, \lambda)$                     & -                                                     & -                                                     & $-0.012793 \pm 0.074834$                              \\
			$(M, \Delta x)$                    & -                                                     & -                                                     & $0.147159 \pm 0.106346$                               \\
			$(M, \lambda)$                     & -                                                     & -                                                     & $0.053237 \pm 0.082949$                               \\
			$(\Delta x, \lambda)$              & -                                                     & -                                                     & $-0.061285 \pm 0.080908$                              \\
			\hline
		\end{tabular}
	\end{center}
\end{table*}

\begin{table*}[b]
	\caption{PSNR: GS-PINN sensitivity indices (ST, S1, S2) for $\textbf{\textup{h}}_{\textup{inner}}.$}
	\label{tab:GSPINN_sa_25_PSNR}
	\begin{center}
		\begin{tabular}{|l|c|c|c|}
			\hline
			$\textbf{\textup{Parameter}}$      & $\textbf{\textup{ST}} \pm \textbf{\textup{ST\_conf}}$ & $\textbf{\textup{S1}} \pm \textbf{\textup{S1\_conf}}$ & $\textbf{\textup{S2}} \pm \textbf{\textup{S2\_conf}}$ \\
			\hline
			$\textup{distance} (d)$            & $0.217436 \pm 0.044753$                               & $0.030065 \pm 0.074212$                               & -                                                     \\
			$\textup{SLM\_size} (M)$           & $0.863566 \pm 0.153941$                               & $0.571657 \pm 0.147347$                               & -                                                     \\
			$\textup{pixel\_pitch} (\Delta x)$ & $0.407003 \pm 0.067504$                               & $-0.003727 \pm 0.097031$                              & -                                                     \\
			$\textup{wavelength} (\lambda)$    & $0.175482 \pm 0.041494$                               & $0.009893 \pm 0.077326$                               & -                                                     \\
			\hline
			$(d, M)$                           & -                                                     & -                                                     & $0.033311 \pm 0.112567$                               \\
			$(d, \Delta x)$                    & -                                                     & -                                                     & $0.071294 \pm 0.116655$                               \\
			$(d, \lambda)$                     & -                                                     & -                                                     & $-0.000381 \pm 0.107515$                              \\
			$(M, \Delta x)$                    & -                                                     & -                                                     & $0.151585 \pm 0.206442$                               \\
			$(M, \lambda)$                     & -                                                     & -                                                     & $-0.088123 \pm 0.175146$                              \\
			$(\Delta x, \lambda)$              & -                                                     & -                                                     & $0.063540 \pm 0.138643$                               \\
			\hline
		\end{tabular}
	\end{center}
\end{table*}

\begin{table*}[b]
	\caption{SSIM: GS-PINN sensitivity indices (ST, S1, S2) for $\textbf{\textup{h}}_{\textup{inner}}.$}
	\label{tab:GSPINN_sa_25_SSIM}
	\begin{center}
		\begin{tabular}{|l|c|c|c|}
			\hline
			$\textbf{\textup{Parameter}}$      & $\textbf{\textup{ST}} \pm \textbf{\textup{ST\_conf}}$ & $\textbf{\textup{S1}} \pm \textbf{\textup{S1\_conf}}$ & $\textbf{\textup{S2}} \pm \textbf{\textup{S2\_conf}}$ \\
			\hline
			$\textup{distance} (d)$            & $0.457315\pm 0.120909$                                & $0.006375 \pm 0.106556$                               & -                                                     \\
			$\textup{SLM\_size} (M)$           & $0.925866 \pm 0.178024$                               & $0.271305 \pm 0.162116$                               & -                                                     \\
			$\textup{pixel\_pitch} (\Delta x)$ & $0.601750 \pm 0.122240$                               & $-0.121822 \pm 0.119976$                              & -                                                     \\
			$\textup{wavelength} (\lambda)$    & $0.411932 \pm 0.109048$                               & $0.012968 \pm 0.100034$                               & -                                                     \\
			\hline
			$(d, M)$                           & -                                                     & -                                                     & $0.023183 \pm 0.141278$                               \\
			$(d, \Delta x)$                    & -                                                     & -                                                     & $0.107732 \pm 0.135216$                               \\
			$(d, \lambda)$                     & -                                                     & -                                                     & $0.069590 \pm 0.148810$                               \\
			$(M, \Delta x)$                    & -                                                     & -                                                     & $0.379565 \pm 0.232710$                               \\
			$(M, \lambda)$                     & -                                                     & -                                                     & $0.076986 \pm 0.202205$                               \\
			$(\Delta x, \lambda)$              & -                                                     & -                                                     & $0.158230 \pm 0.186476$                               \\
			\hline
		\end{tabular}
	\end{center}
\end{table*}

\begin{table*}[b]
	\caption{PSNR: GS-PINN sensitivity indices (ST, S1, S2) for $\textbf{\textup{h}}_{\textup{outer}}.$}
	\label{tab:GSPINN_sa_75_PSNR}
	\begin{center}
		\begin{tabular}{|l|c|c|c|}
			\hline
			$\textbf{\textup{Parameter}}$      & $\textbf{\textup{ST}} \pm \textbf{\textup{ST\_conf}}$ & $\textbf{\textup{S1}} \pm \textbf{\textup{S1\_conf}}$ & $\textbf{\textup{S2}} \pm \textbf{\textup{S2\_conf}}$ \\
			\hline
			$\textup{distance} (d)$            & $0.056764 \pm 0.010912$                               & $-0.000460 \pm 0.038566$                              & -                                                     \\
			$\textup{SLM\_size} (M)$           & $0.835761 \pm 0.124853$                               & $0.734959 \pm 0.119901$                               & -                                                     \\
			$\textup{pixel\_pitch} (\Delta x)$ & $0.096799 \pm 0.017366$                               & $0.024120 \pm 0.054834$                               & -                                                     \\
			$\textup{wavelength} (\lambda)$    & $0.080277 \pm 0.013127$                               & $-0.001338 \pm 0.046227$                              & -                                                     \\
			\hline
			$(d, M)$                           & -                                                     & -                                                     & $0.011516 \pm 0.048397$                               \\
			$(d, \Delta x)$                    & -                                                     & -                                                     & $0.014439 \pm 0.052719$                               \\
			$(d, \lambda)$                     & -                                                     & -                                                     & $0.022735 \pm 0.056129$                               \\
			$(M, \Delta x)$                    & -                                                     & -                                                     & $0.048144 \pm 0.095641$                               \\
			$(M, \lambda)$                     & -                                                     & -                                                     & $0.051504 \pm 0.075261$                               \\
			$(\Delta x, \lambda)$              & -                                                     & -                                                     & $-0.027780 \pm 0.090817$                              \\
			\hline
		\end{tabular}
	\end{center}
\end{table*}

\begin{table*}[b]
	\caption{SSIM: GS-PINN sensitivity indices (ST, S1, S2) for $\textbf{\textup{h}}_{\textup{outer}}.$}
	\label{tab:GSPINN_sa_75_SSIM}
	\begin{center}
		\begin{tabular}{|l|c|c|c|}
			\hline
			$\textbf{\textup{Parameter}}$      & $\textbf{\textup{ST}} \pm \textbf{\textup{ST\_conf}}$ & $\textbf{\textup{S1}} \pm \textbf{\textup{S1\_conf}}$ & $\textbf{\textup{S2}} \pm \textbf{\textup{S2\_conf}}$ \\
			\hline
			$\textup{distance} (d)$            & $0.238366\pm 0.053205$                                & $-0.054770  \pm 0.090144$                             & -                                                     \\
			$\textup{SLM\_size} (M)$           & $0.862056 \pm 0.129844$                               & $0.676792 \pm 0.138666$                               & -                                                     \\
			$\textup{pixel\_pitch} (\Delta x)$ & $0.203247 \pm 0.050173$                               & $-0.017016 \pm 0.079095$                              & -                                                     \\
			$\textup{wavelength} (\lambda)$    & $0.181209 \pm 0.043596$                               & $-0.049422 \pm 0.078540$                              & -                                                     \\
			\hline
			$(d, M)$                           & -                                                     & -                                                     & $0.063584 \pm 0.108349$                               \\
			$(d, \Delta x)$                    & -                                                     & -                                                     & $0.055035 \pm 0.103875$                               \\
			$(d, \lambda)$                     & -                                                     & -                                                     & $0.088502 \pm 0.107233$                               \\
			$(M, \Delta x)$                    & -                                                     & -                                                     & $0.045500 \pm 0.107128$                               \\
			$(M, \lambda)$                     & -                                                     & -                                                     & $0.054164 \pm 0.115869$                               \\
			$(\Delta x, \lambda)$              & -                                                     & -                                                     & $-0.009580 \pm 0.102762$                              \\
			\hline
		\end{tabular}
	\end{center}
\end{table*}


\begin{table*}[b]
	\caption{PSNR: GS algorithm sensitivity indices (ST, S1) for $\textbf{\textup{h}}_{\textup{mid}}.$}
	\label{tab:GS_sa_50_PSNR}
	\begin{center}
		\begin{tabular}{|l|l|c|c|}
			\hline
			$\textbf{\textup{Iteration}}$ & $\textbf{\textup{Parameter}}$      & $\textbf{\textup{ST}} \pm \textbf{\textup{ST\_conf}}$ & $\textbf{\textup{S1}} \pm \textbf{\textup{S1\_conf}}$ \\
			\hline
			$1$                           & $\textup{distance} (d)$            & $0.418797\pm 0.070810$                                & $-0.0095605 \pm 0.057152$                             \\
			$1$                           & $\textup{SLM\_size} (M)$           & $0.532494 \pm 0.071666$                               & $0.009924 \pm 0.070842$                               \\
			$1$                           & $\textup{pixel\_pitch} (\Delta x)$ & $0.89645 \pm 0.085240$                                & $0.279964 \pm 0.098191$                               \\
			$1$                           & $\textup{wavelength} (\lambda)$    & $0.32252 \pm 0.080749$                                & $0.013781 \pm 0.050062$                               \\
			\hline
			$30$                          & $\textup{distance} (d)$            & $0.575011\pm 0.06580$                                 & $0.073236 \pm 0.064654$                               \\
			$30$                          & $\textup{SLM\_size} (M)$           & $0.554901 \pm 0.050661$                               & $0.016904 \pm 0.063953$                               \\
			$30$                          & $\textup{pixel\_pitch} (\Delta x)$ & $0.79172 \pm 0.069831$                                & $0.122615 \pm 0.084754$                               \\
			$30$                          & $\textup{wavelength} (\lambda)$    & $0.28729 \pm 0.047131$                                & $0.00006 \pm 0.041671$                                \\
			\hline
		\end{tabular}
	\end{center}
\end{table*}

\begin{table*}[b]
	\caption{SSIM: GS algorithm sensitivity indices (ST, S1) for $\textbf{\textup{h}}_{\textup{mid}}.$}
	\label{tab:GS_sa_50_SSIM}
	\begin{center}
		\begin{tabular}{|l|l|c|c|}
			\hline
			$\textbf{\textup{Iteration}}$ & $\textbf{\textup{Parameter}}$      & $\textbf{\textup{ST}} \pm \textbf{\textup{ST\_conf}}$ & $\textbf{\textup{S1}} \pm \textbf{\textup{S1\_conf}}$ \\
			\hline
			$1$                           & $\textup{distance} (d)$            & $0.478523\pm 0.0900421$                               & $0.014104 \pm 0.071935$                               \\
			$1$                           & $\textup{SLM\_size} (M)$           & $0.513308 \pm 0.056194$                               & $0.070332 \pm 0.083237$                               \\
			$1$                           & $\textup{pixel\_pitch} (\Delta x)$ & $0.86733 \pm 0.109440$                                & $0.113502 \pm 0.089765$                               \\
			$1$                           & $\textup{wavelength} (\lambda)$    & $0.282547 \pm 0.059899$                               & $-0.012099 \pm 0.036987$                              \\
			\hline
			$30$                          & $\textup{distance} (d)$            & $0.376298\pm 0.067010$                                & $-0.0057486 \pm 0.052238$                             \\
			$30$                          & $\textup{SLM\_size} (M)$           & $0.459715 \pm 0.054766$                               & $0.0406021 \pm 0.064098$                              \\
			$30$                          & $\textup{pixel\_pitch} (\Delta x)$ & $0.904809 \pm 0.085998$                               & $0.3068320 \pm 0.0965937$                             \\
			$30$                          & $\textup{wavelength} (\lambda)$    & $0.284520 \pm 0.0561640$                              & $0.004076 \pm 0.035673$                               \\
			\hline
		\end{tabular}
	\end{center}
\end{table*}

\begin{table*}[b]
	\caption{PSNR: GS algorithm sensitivity indices (ST, S1) for $\textbf{\textup{h}}_{\textup{inner}}$ and $\textbf{\textup{h}}_{\textup{outer}}.$}
	\label{tab:GS_sa_25_75_PSNR}
	\begin{center}
		\begin{tabular}{|l|l|c|c|}
			\hline
			$\textbf{\textup{Iteration}}$ & $\textbf{\textup{Parameter}}$      & $\textbf{\textup{ST}} \pm \textbf{\textup{ST\_conf}} (\textbf{\textup{h}}_{\textup{inner}}.)$ & $\textbf{\textup{ST}} \pm \textbf{\textup{ST\_conf}} (\textbf{\textup{h}}_{\textup{outer}}.)$ \\
			\hline
			$1$                           & $\textup{distance} (d)$            & $0.506241\pm 0.129022$                                                               & $0.1872011 \pm 0.039086$                                                             \\
			$1$                           & $\textup{SLM\_size} (M)$           & $0.420179 \pm 0.097640$                                                              & $0.344047 \pm 0.0613662$                                                             \\
			$1$                           & $\textup{pixel\_pitch} (\Delta x)$ & $0.883636 \pm 0.147613$                                                              & $0.426114 \pm 0.076417$                                                              \\
			$1$                           & $\textup{wavelength} (\lambda)$    & $0.320455 \pm 0.101777$                                                              & $0.072722 \pm 0.0138883$                                                             \\
			\hline
			$30$                          & $\textup{distance} (d)$            & $0.678911\pm 0.1230092$                                                              & $0.220039 \pm 0.045422$                                                              \\
			$30$                          & $\textup{SLM\_size} (M)$           & $0.478841 \pm 0.110541$                                                              & $0.1788105 \pm 0.035432$                                                             \\
			$30$                          & $\textup{pixel\_pitch} (\Delta x)$ & $0.850293 \pm 0.138281$                                                              & $0.509250 \pm 0.088038$                                                              \\
			$30$                          & $\textup{wavelength} (\lambda)$    & $0.325285 \pm 0.100448$                                                              & $0.078978 \pm 0.017330$                                                              \\
			\hline
		\end{tabular}
	\end{center}
\end{table*}

\begin{table*}[b]
	\caption{SSIM: GS algorithm sensitivity indices (ST, S1) for $\textbf{\textup{h}}_{\textup{inner}}$ and $\textbf{\textup{h}}_{\textup{outer}}.$}
	\label{tab:GS_sa_25_75_SSIM}
	\begin{center}
		\begin{tabular}{|l|l|c|c|}
			\hline
			$\textbf{\textup{Iteration}}$ & $\textbf{\textup{Parameter}}$      & $\textbf{\textup{ST}} \pm \textbf{\textup{ST\_conf}} (\textbf{\textup{h}}_{\textup{inner}}.)$ & $\textbf{\textup{ST}} \pm \textbf{\textup{ST\_conf}} (\textbf{\textup{h}}_{\textup{outer}}.)$ \\
			\hline
			$1$                           & $\textup{distance} (d)$            & $0.548783\pm 0.1470908$                                                              & $0.115974 \pm 0.027301$                                                              \\
			$1$                           & $\textup{SLM\_size} (M)$           & $0.386270 \pm 0.094665$                                                              & $0.852533 \pm 0.116372$                                                              \\
			$1$                           & $\textup{pixel\_pitch} (\Delta x)$ & $0.725548 \pm 0.1284623$                                                             & $0.121418 \pm 0.025218$                                                              \\
			$1$                           & $\textup{wavelength} (\lambda)$    & $0.275962 \pm 0.085992$                                                              & $0.1483183 \pm 0.0317105$                                                            \\
			\hline
			$30$                          & $\textup{distance} (d)$            & $0.4701971\pm 0.1275617$                                                             & $0.1020604 \pm 0.0225182$                                                            \\
			$30$                          & $\textup{SLM\_size} (M)$           & $0.349402 \pm 0.096186$                                                              & $0.4653550 \pm 0.0838963$                                                            \\
			$30$                          & $\textup{pixel\_pitch} (\Delta x)$ & $0.823904 \pm 0.1366968$                                                             & $0.3583915 \pm 0.064410$                                                             \\
			$30$                          & $\textup{wavelength} (\lambda)$    & $0.266439 \pm 0.0931318$                                                             & $0.0766469 \pm 0.0154371$                                                            \\
			\hline
		\end{tabular}
	\end{center}
\end{table*}
\clearpage
\subsection{Visualization for h\_mid FMH for different FMs.}
\label{supplemetary:sec:D}
\subsubsection{Forward model Sensitivity -  Caveats}
\begin{figure}[H]
	\centering
        \includegraphics[width=3.5in]{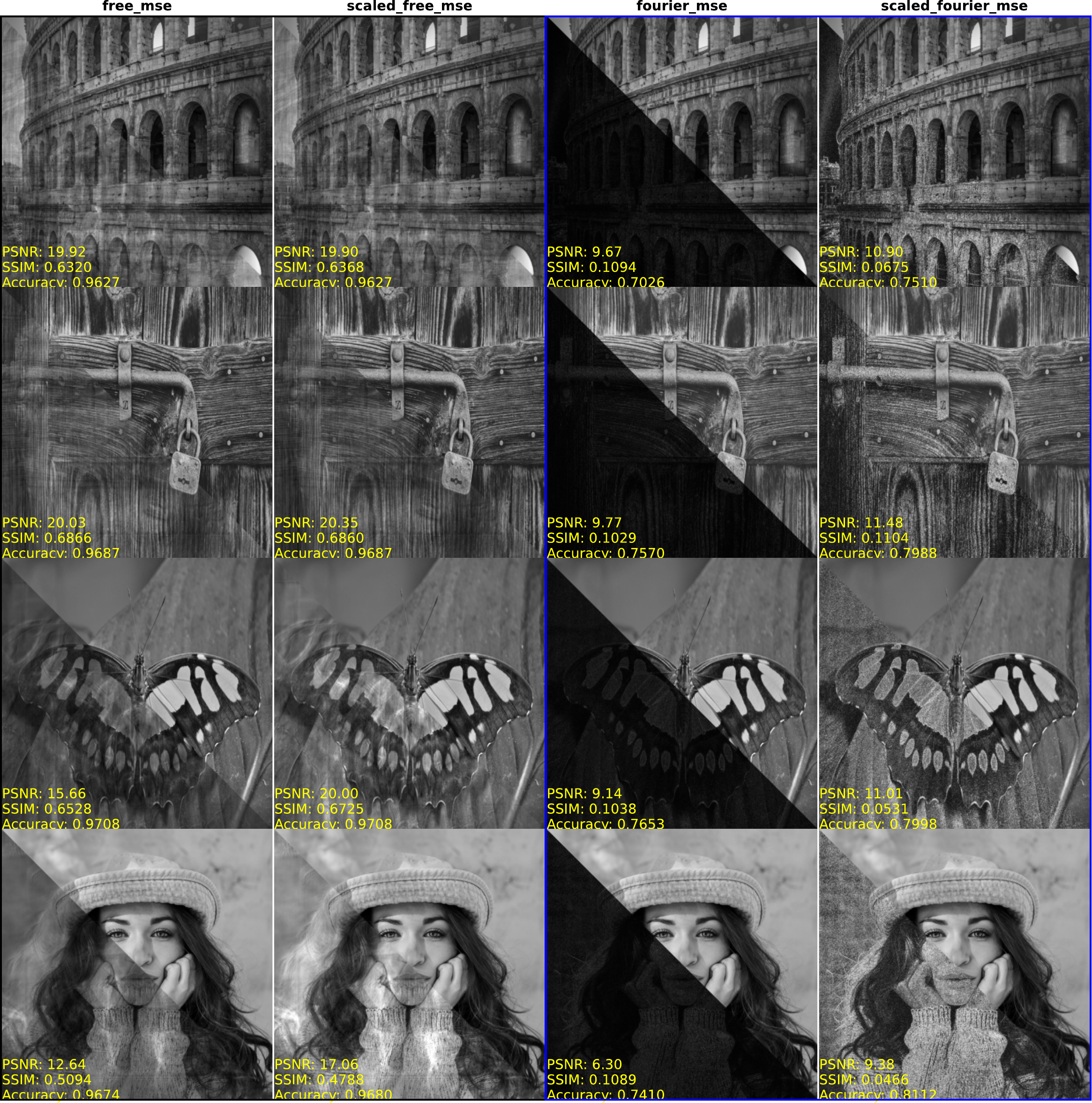}
	\caption{Visualization of network performance using the mean squared error (MSE) loss function for $\textbf{h}_{\textbf{mid}}$ FMH across different forward models. The first two columns (black-bordered) correspond to free space propagation, while the last two columns (blue-bordered) represent Fourier holography. The upper triangular region in each panel shows the original image, while the lower triangular region displays the GS-PINN output. To ensure comparability, the outputs in the second and fourth columns are scaled to match the mean intensity of the corresponding original images. Performance metrics-including PSNR, SSIM, and accuracy $\left(A = \frac{{\sum \widetilde{I}({x,y,z} )\; \; I({x,y,z} )}}{{ \sqrt{\sum I({x,y,z} )^{2}
    \sum \widetilde{I}({x,y,z} )^{2}}}}\right)$ are highlighted in yellow.}
	\label{fig:0010_h_mid_nn_mse_visualization}
\end{figure}

\begin{figure}[H]
	\centering
        \includegraphics[width=3.5in]{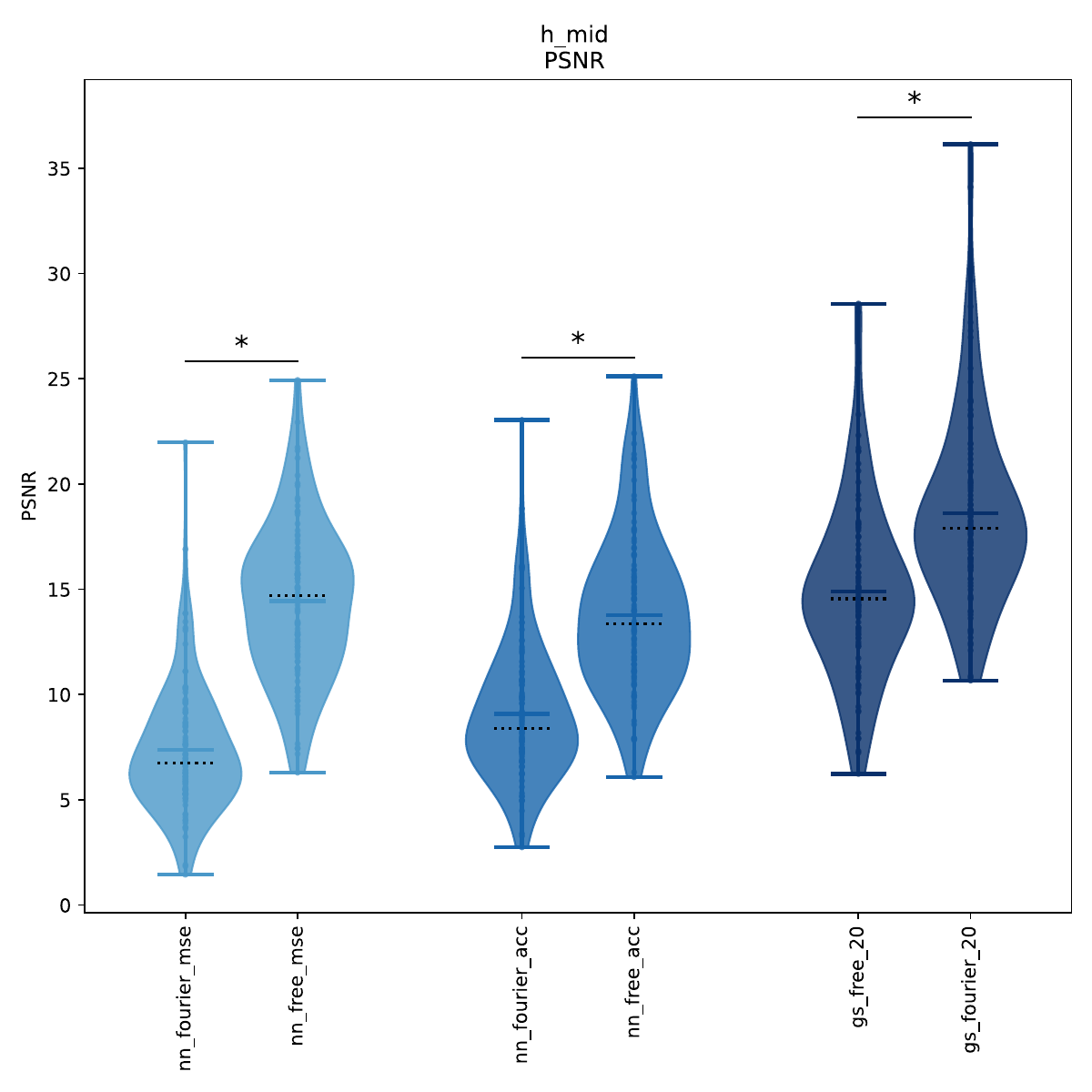}
	\caption
    {Comparison of different forward models (FMs) for GS-PINN using mean squared error (MSE) and accuracy as loss functions, alongside the GS algorithm for $\textbf{h}_{\textbf{mid}}$ FMH. Models are labelled as X\_Y\_Z, where X denotes either GS-PINN (nn) or the GS algorithm (gs), Y specifies the FM (free for free space propagation or fourier for Fourier holography), and Z indicates the loss function (MSE or Accuracy) or the number of iterations (20). Violin plots (with medians shown as dotted lines, and extreme and mean values indicated) illustrate the comparative performance of the two FMs for both GS-PINN and the GS algorithm. A Wilcoxon signed-rank test (one-sided, alternative hypothesis: ``less", n=100) confirmed a statistically significant difference (p $<$ 0.025, marked *), showing that free-space propagation outperforms Fourier holography for GS-PINN trained with different loss functions, whereas the opposite trend is observed for the GS algorithm.}
	\label{fig:0010_h_mid_violin_visualization}
\end{figure}

\begin{figure*}
	\centering
        \includegraphics[width=4.5in]{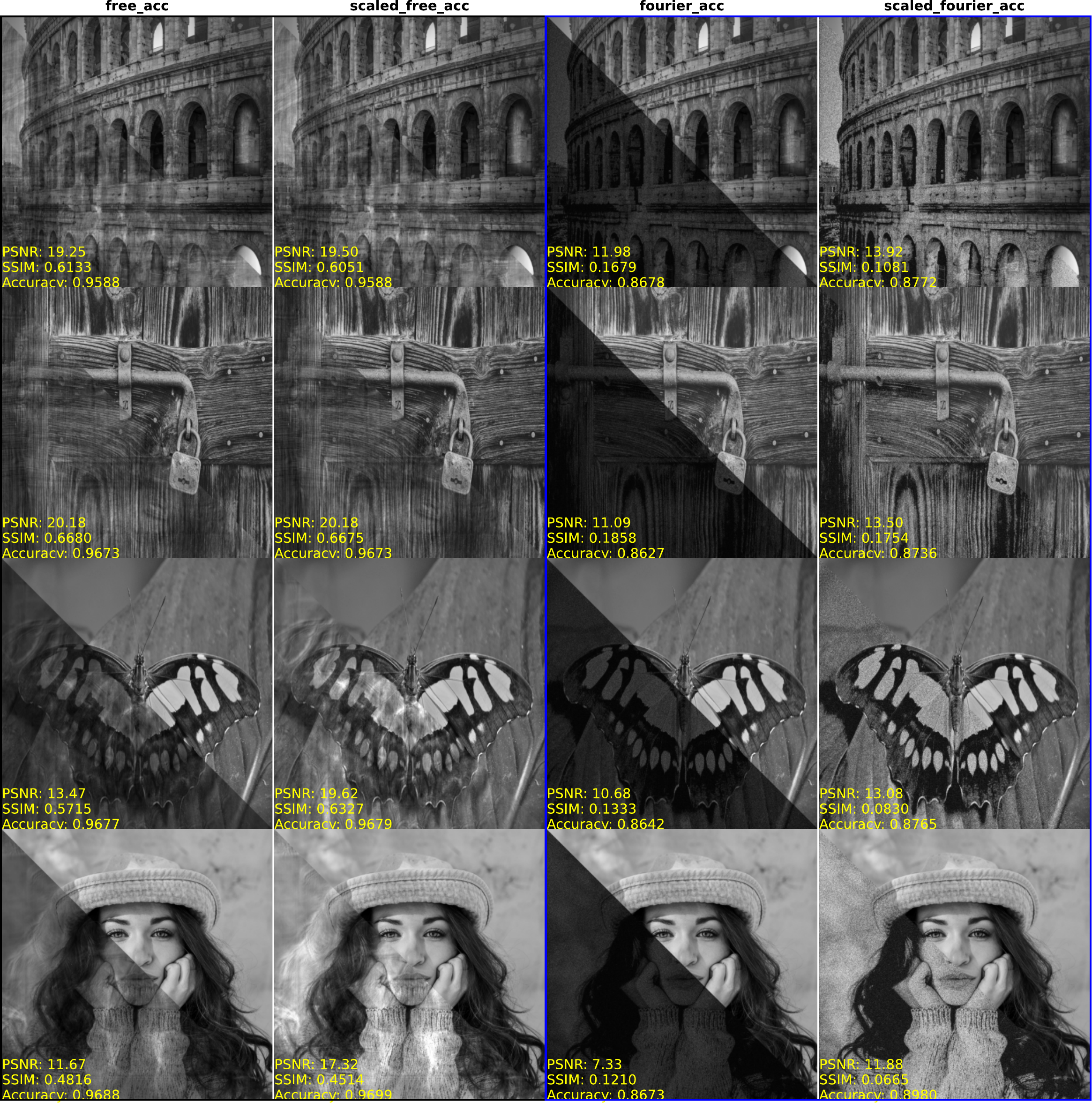}
	\caption{Visualization of network performance using the accuracy loss function $\left(A = \frac{{\sum \widetilde{I}({x,y,z} )\; \; I({x,y,z} )}}{{ \sqrt{\sum I({x,y,z} )^{2}
    \sum \widetilde{I}({x,y,z} )^{2}}}}\right)$ for $\textbf{h}_{\textbf{mid}}$ FMH across different forward models. The first two columns (black-bordered) correspond to free space propagation, while the last two columns (blue-bordered) represent Fourier holography. The upper triangular region in each panel shows the original image, while the lower triangular region displays the GS-PINN output. To ensure comparability, the outputs in the second and fourth columns are scaled to match the mean intensity of the corresponding original images. Performance metrics-including PSNR, SSIM, and accuracy are highlighted in yellow.}
	\label{fig:0010_h_mid_nn_acc_visualization}
\end{figure*}

\begin{figure*}
	\centering
        \includegraphics[width=4.5in]{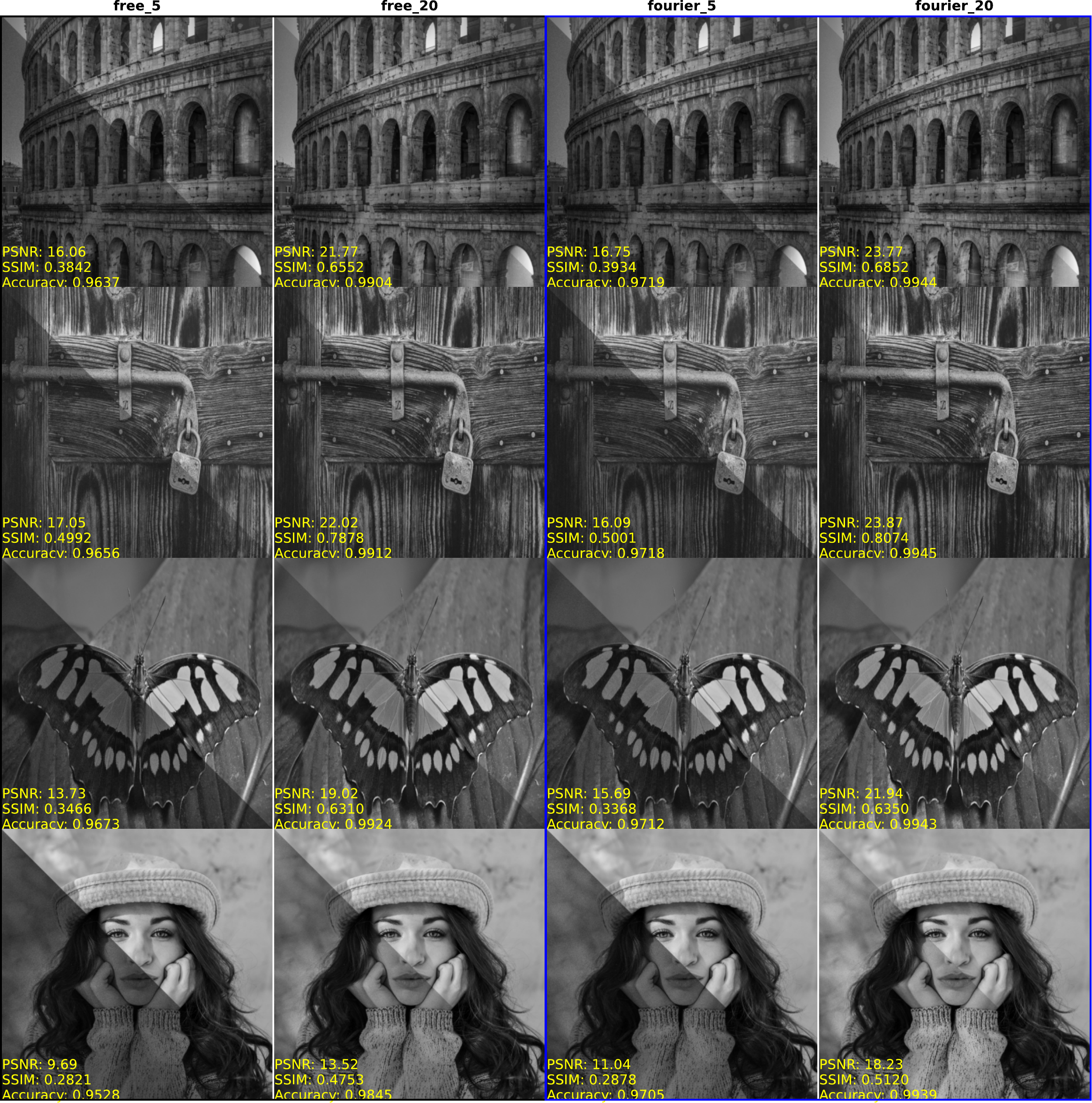}
	\caption{Visualization of GS algorithm performance for $\textbf{h}_{\textbf{mid}}$ FMH across different forward models. The first two columns (black-bordered) correspond to free space propagation, while the last two columns (blue-bordered) represent Fourier holography. The upper triangular region in each panel shows the original image, while the lower triangular region displays the GS algorithm output. Performance metrics-including PSNR, SSIM, and accuracy are highlighted in yellow for iterations 5 and 20 for both the FMs.}
	\label{fig:0010_h_mid_gs_visualization}
\end{figure*}

For the $\textbf{\textup{h}}_{\textup{mid}}$ configuration, in the GS algorithm, the mean intensity is better approximated for both forward models as the number of iterations increases \figref{fig:0010_h_mid_gs_visualization}. For the base GS-PINN models trained on the $\textbf{\textup{h}}_{\textup{mid}}$ configuration, the $\texttt{base\_free}$ model outperforms the $\texttt{base\_fourier}$ model for different loss functions \figref{fig:0010_h_mid_violin_visualization}. Notably, the $\texttt{base\_fourier}$ GS-PINN network struggles to accurately approximate the average DC component for both loss functions. After scaling the hologram intensity to match the mean intensity of the original image, the hologram features are preserved, indicating that the primary limitation lies in mean intensity estimation rather than feature representation \figref{fig:0010_h_mid_nn_mse_visualization} - \figref{fig:0010_h_mid_nn_acc_visualization}. While the $\texttt{base\_free}$ model initially performs better than the $\texttt{base\_fourier}$ model, fine-tuning with different forward models results in a performance ranking dictated by the forward model rather than the base model. Specifically, the superior performance of $\texttt{base\_fourier\_free}$ over both $\texttt{base\_fourier\_fourier}$ and $\texttt{base\_free\_fourier}$ suggests that free-space propagation improves model performance, regardless of the initial base model \figref{fig:gspinn_free_better_four_violin_psnr} - \figref{fig:gspinn_free_better_four_violin_ssim}. This finding underscores the crucial role of the forward model in determining the final performance, effectively isolating its impact from differences in baseline models.

\clearpage

\end{document}